\documentclass[]{cit_lab_mfr}

\usepackage{amssymb}

\usepackage{amsthm}
\usepackage{caption}
\usepackage{amsmath}
\usepackage{makecell}
\usepackage{mathtools}
\usepackage{enumitem}
\usepackage{hyperref}
\usepackage{cleveref}
\usepackage{verbatim}
\usepackage{wrapfig}  
\usepackage{graphicx}
\usepackage{floatrow}
\usepackage{subcaption}
\usepackage{listings}
\usepackage{algorithm}
\usepackage{microtype}
\usepackage{graphicx}
\usepackage{subcaption}
\usepackage{booktabs} 
\usepackage{multirow}
\usepackage{graphicx}
\usepackage[normalem]{ulem}
\usepackage{enumitem}
\useunder{\uline}{\ul}{}
\usepackage[utf8]{inputenc}
\usepackage{xcolor}
\usepackage{graphicx}

\usepackage{wrapfig}
\usepackage{graphicx}
\usepackage{needspace}

\definecolor{DarkGreen}{HTML}{006400}
\definecolor{DarkRed}{HTML}{8B0000}
\definecolor{diy_pink}{RGB}{255,247,240}

\usepackage{multirow}         
\usepackage[table]{xcolor}     

\usepackage{graphicx}   
\usepackage{adjustbox}  
\usepackage{array}      
\usepackage{graphicx}
\usepackage{caption} 
\usepackage{xcolor}
\usepackage{pifont}

\usepackage{hyperref}

\usepackage{subcaption} %

\usepackage[toc,page,header]{appendix}

\usepackage{minitoc}

\usepackage{amsmath}

\usepackage{mathtools}

\usepackage{xurl}

\RequirePackage{algorithmic}

\title{OmniFysics-Nano-V2 Technical Report: \\ Understanding the Physical World Across Modalities}

\author{%
\parbox{\textwidth}{\centering
Yizhou Liu$^{*}$,
Jinghang Han$^{*}$,
Kaixiang Qiu,
Qi He,
Minghao Han,
Yue Jiang,
Xujia Chen, \\
Wei Zou,
Shunli Wang$^{\S}$,
Lihua Zhang$^{\S}$,
Dingkang Yang$^{\dagger,\S}$
}}

\affiliation{%
\parbox{\textwidth}{\centering\small
Physical Superintelligence Lab, Fysics AI \\[1mm]
College of Intelligent Robotics and Advanced Manufacturing, Fudan University\\[1mm]
}}

\contribution[*]{Equal contribution}
\contribution[\dagger]{Project lead}
\contribution[\S]{Corresponding author}

\abstract{
Omni-modal models have expanded multimodal interaction across vision, audio, speech, and language. However, their training is predominantly organized around semantic descriptions and general-purpose objectives, leaving physical attributes, interaction states, and causal mechanisms only partially specified. This gap is not simply a matter of modality coverage: adding more modalities does not by itself provide the supervision needed to connect observations with the physical structure of the world. We present OmniFysics-Nano-V2, a compact omni-modal model for physical-world perception and understanding. The model supports image, video, audio, speech, and text inputs within a shared reasoning framework, together with text and speech generation. To address the lack of explicit physical supervision, we construct a dual-branch physics-aware data pipeline that grounds salient objects in structured physical attributes and aligns visual changes with acoustic events, intermediate responses, and interaction outcomes. To address homogeneous training objectives, we curate reinforcement-learning prompts by reward diversity and adopt a two-stage Group Relative Policy Optimization curriculum that progresses from general task correctness to fine-grained physical perceptual reasoning. 
Experiments across multimodal, audio-visual, and physical reasoning benchmarks show that the proposed data and training strategy improves physical-world understanding while preserving broad omni-modal competence. 
The proposed model achieves leading result on 17 of 21 benchmarks against SOTA omni-modal models.
By equipping AI systems with both omni-modal and physical-world perception capabilities, OmniFysics-Nano-V2 is poised to become a cornerstone of next-generation Physical AI.
}
\date{\today}
\checkdata[Corresponding]{ \url{dicken@fyscis.ai}, \url{lihuazhang@fudan.edu.cn}}
\checkdata[Page]{\url{https://github.com/Fysics-AI/OmniFysics-Nano-V2}}
\checkdata[Hugging Face]{\url{https://huggingface.co/Fysics-AI/OmniFysics-Nano-V2}}

\begin{document}
\maketitle

\section{Introduction}
Understanding the physical world requires more than recognizing objects, understanding scenes, or following language instructions.
Currently, most Multimodal Large Language Models (MLLMs) \cite{bai2025qwen25vltechnicalreport, li2024llavaonevisioneasyvisualtask} focus primarily on vision‑language interaction in general‑purpose scenarios.
Despite the promising performance of these models on visual recognition, Visual Question Answering (VQA), and instruction following~\cite{bai2025qwen3vl,wang2025internvl35advancingopensourcemultimodal,an2026llavaonevision2nextgenerationperceptualintelligence}, there remains a fundamental gap compared with physical-world understanding.
A model endowed with physical‑reasoning capabilities should resemble a human agent equipped with rich physical prior knowledge.
It can infer the static physical properties of real-world objects, as well as comprehend dynamic physical events observed in video inputs.
Physical perception and omni-modal understanding constitute indispensable foundations for the next generation of artificial intelligence ~\cite{han2026omnifysicsphysicalintelligenceevolution}.

\begin{figure}[t]
    \centering
    \includegraphics[width=0.65\linewidth]{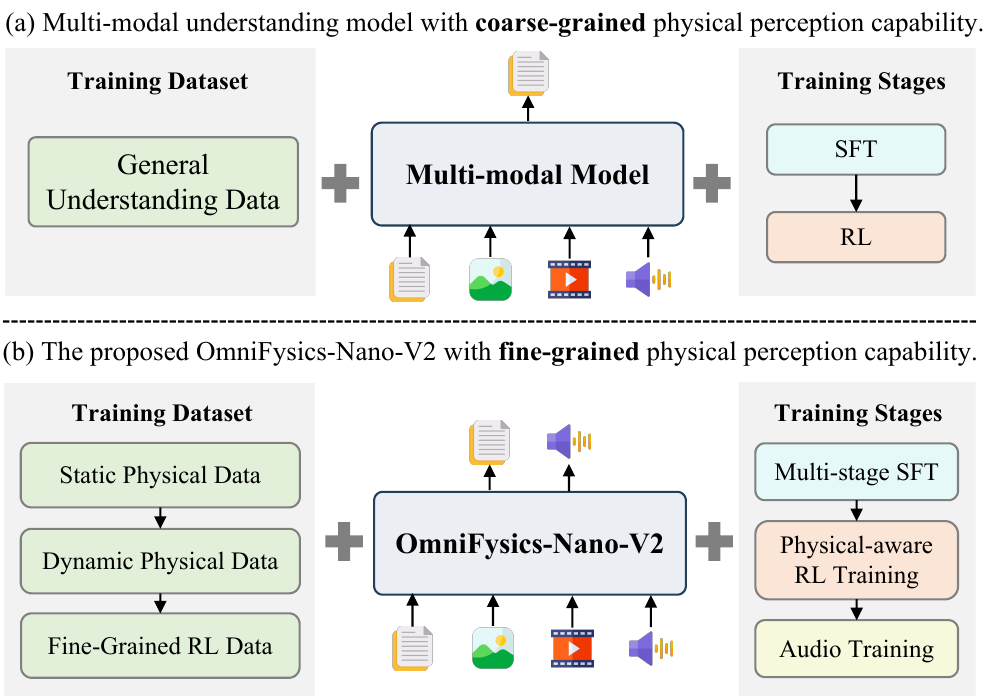}
    \caption{Illustration of general multimodal models and the proposed OmniFysics-Nano-V2. In contrast to generic datasets and training strategies, we design a physics-aware data pipeline that incorporates both static-property and dynamic-event data. Moreover, we adopt multi-stage reinforcement learning to boost the model’s physical perception capabilities.}
    \label{fig1}
\end{figure}

\begin{figure*}[t]
    \centering
    \includegraphics[width=1.0\linewidth]{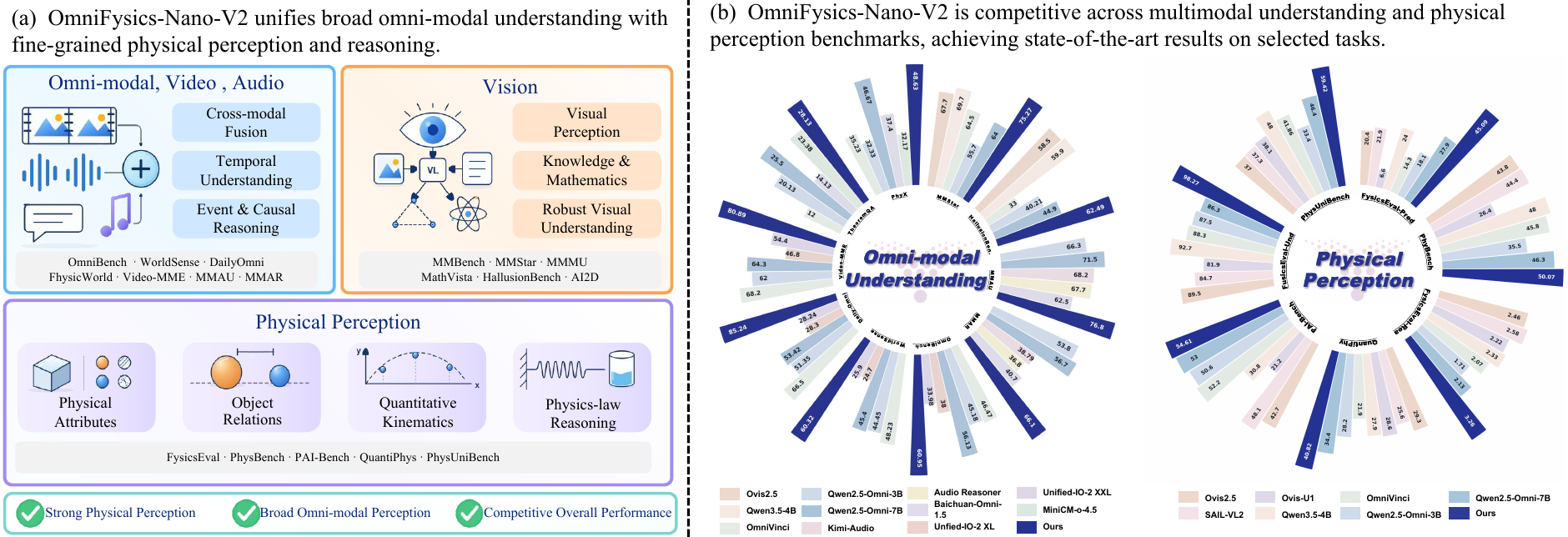}
    \caption{Demonstration of OmniFysics-Nano-V2 model capability and performance. The proposed physics-aware data pipeline and efficient policy-aware training strategies endow the model with a powerful ability to understand the physical world.}
    \label{fig2}
\end{figure*}

Several existing omni‑modal models have gone beyond the conventional vision‑language interaction paradigm and are capable of performing understanding tasks under interleaved multi‑modal information.
For instance, closed‑source models include GPT‑4o ~\cite{openai2024gpt4o} and Gemini~\cite{geminiteam2023gemini}, while open‑source ones cover Qwen2.5-Omni~\cite{xu2025qwen25omni}, Qwen3-Omni~\cite{xu2025qwen3omni}, OmniVinci~\cite{ye2025omnivinci} and MiniCPM-o-4.5~\cite{cui2026minicpmo45}.
Existing omni-modal systems are generally trained on web-scale semantic data and optimized for transcription, captioning, instruction following.
They therefore learn to align modalities at the level of meaning, but receive little explicit supervisory information to identify specific physical properties, temporally organize an interaction, or distinguish a physically grounded explanation from a plausible description.
These studies fully reveal the complementary nature of information across different modalities. 
For example, visual signals enable the inference of object geometry and motion, while audio information allows more precise localization of event timings.
However, merely adding more modalities cannot directly endow models with physical understanding capabilities.

We argue that two fundamental factors currently limit the physical understanding capabilities of omni-modal models: \textbf{Absent Physical Supervision} and \textbf{Homogeneous Training Strategies}.
At the level of dataset and supervisory information, generic multi-modal datasets~\cite{ST-VQA, AV-Odyssey, Video-MME-v2} predominantly provide semantic descriptions of objects and events, while physical attributes, interaction states, and causal mechanisms are often implicit, ambiguous, or completely absent.
Therefore, models struggle to associate cross-modal observations with the latent physical properties and laws governing real-world interactions.
At the level of training strategy, existing models \cite{xu2025qwen25omni,xu2025qwen3omni,bai2025qwen3vl} are typically trained using general-purpose and largely homogeneous objectives.
These objectives tend to prioritize general semantic understanding and cross-modal alignment over the explicit modeling of physical dynamics and causal relationships, thereby encouraging models to exploit superficial semantic correlations rather than acquire robust physical reasoning abilities.

To remedy \textit{Absent Physical Supervision}, we build a physics-aware data pipeline with complementary \textbf{Static} and \textbf{Dynamic} branches.
The static branch anchors salient objects to a structured physical prototype bank, augments them with intrinsic-property annotations, and enforces plausible value ranges and cross-property compatibility.
Rather than treating frames in isolation, the dynamic branch focuses on event-centered video clips, aligning visible motion and state transitions with acoustic transients to trace an interaction from its initial state, through intermediate responses, to its final outcome.
Together, the two branches connect object identity and intrinsic physical properties to the visual and acoustic evidence through which those properties are revealed during interaction.

At the training stage, we move beyond \textit{Homogeneous Training Strategies} with an efficient, policy-aware two-stage curriculum.
Before optimization, we sample four stochastic rollouts for each candidate prompt and compare their rewards. Only prompts that produce reward variation are retained, as they provide meaningful within-group learning signals.
Applied to mathematical, image–text, video, and mixed-modal reinforcement learning data, this reward-diversity filter removes over half of the candidate samples and saves more than 2,600 GPU-hours.
We then organize GRPO into a two-stage curriculum: the first stage improves answer accuracy on general tasks, while the second further strengthens physical understanding through format-specific rewards.
By combining selective data curation with curriculum-based optimization, the resulting policy learns not only to make accurate predictions but also to support its decisions with physically meaningful evidence.

In this paper, we propose \textbf{OmniFysics-Nano-V2}, a compact 4B omni-modal model for physical-world perception and understanding.
Figure \ref{fig1} illustrates the differences between general multimodal models and our model.
The model supports the understanding of image, video, audio, speech, and text inputs, alongside text and audio generation capabilities.
With the help of physics-aware data pipeline and efficient two-stage policy-aware training strategy, our model achieves superior performance across multi-modal perception and physical perception tasks.
Figure~\ref{fig2}(a) illustrates the model’s capabilities in omni-modal data processing and physical understanding in detail.
We evaluated on 21 benchmarks that cover general multimodal, audio, omni-modal / video, physical understanding, mathematical reasoning and physical reasoning benchmarks.
Our model achieves SOTA performance on 17 benchmarks with 4B model size, even against 7B-scale baseline models.
Figure \ref{fig2}(b) demonstrates the superior performance of OmniFysics-Nano-V2 on the Omni-modal Understanding Benchmarks and Physical Perception Benchmarks.
Notably, our model achieves 98.27\% on FysicsEval Understanding~\cite{han2026omnifysicsphysicalintelligenceevolution} and 59.42\% on PhysUniBench~\cite{wang2025physunibench}, surpassing the state-of-the-art baselines by 5.57\% and 11.42\%, respectively.

Our contributions are summarized as follows:

\begin{itemize}
    \item We present \textbf{OmniFysics-Nano-V2}, a compact 4B omni-modal model that supports holistic understanding of images, videos, audios, speeches, and texts with speech output capability.
    \item We design a dual-branch physical supervision pipeline that complements static physical attribute grounding and dynamic physical event modeling.
    \item We develop a reward-diversity filtering for policy-aware RL data curation, and a two-stage GRPO strategy that optimizes general task performance and enhances fine-grained physical reasoning via intermediate perception supervision.
    \item The proposed model achieves leading result on 17 of 21 benchmarks against SOTA omni-modal models, with ablations showing substantial improvements over SFT and large reductions in RL data and compute.
\end{itemize}

\section{Related Work}

\subsection{Omni-Modal Foundation Models}

Contemporary omni-modal foundation models seek to place text, vision, audio, and speech within a single perceptual and generative system. GPT-4o~\cite{openai2024gpt4o} demonstrated end-to-end processing of text, image, audio, and video inputs together with text, audio, and image generation, establishing low-latency speech interaction as a central capability of native multimodal modeling. The Qwen-Omni family develops this direction through time-aligned audiovisual encoding and a Thinker--Talker architecture. Qwen2.5-Omni~\cite{xu2025qwen25omni} supports streaming text and speech generation, Qwen3-Omni~\cite{xu2025qwen3omni} improves long-context perception and modality balance, and Qwen3.5-Omni~\cite{team2026qwen3.5omni} further scales audiovisual training, context length, temporal grounding, and multilingual speech generation. Open models have pursued comparable breadth from different architectural perspectives. Baichuan-Omni-1.5~\cite{li2025baichuan} combines separate visual and acoustic encoders with joint multimodal alignment, OmniVinci~\cite{ye2025omnivinci} strengthens audiovisual fusion through shared latent alignment and explicit temporal encoding, and MiniCPM-o 4.5~\cite{cui2026minicpmo45} organizes perception and speech generation on a common temporal axis to support full-duplex interaction. These systems have substantially advanced modality coverage, streaming response, and cross-modal instruction following, but their notion of unification remains largely functional. Training objectives and evaluations are designed to determine whether heterogeneous signals can be understood and whether coherent responses can be produced across modalities. They seldom require the model to recover latent physical quantities, maintain an explicit account of how object states evolve, or identify the physical event that causally links a visual change with an acoustic observation. Strong omni-modal interaction does not yet amount to a structured model of the physical world.

\subsection{Physical Data Generation and Supervision}

Physical data are distinguished from general multimodal data by the constraints their annotations place on the state or transition underlying an observation. A visual description records what appears in a scene, whereas physical supervision relates that appearance to intrinsic properties, initial conditions, interactions, intermediate responses, or outcomes. Controlled simulation has been the principal source of such supervision because it exposes variables that are difficult to measure in ordinary video. IntPhys~\cite{riochet2020intphysframeworkbenchmarkvisual} evaluates violations of object permanence and continuity, PHYRE~\cite{2019bakhtinPHYRE} tests goal-directed intervention, CLEVRER~\cite{yi2020clevrercollisioneventsvideo} supports causal and counterfactual reasoning over collisions, and Physion~\cite{bear2022physionevaluatingphysicalprediction} makes prediction depend on latent properties including mass, friction, elasticity, and deformability. Procedural platforms such as Kubric~\cite{greff2022kubric} and PhysInOne~\cite{zhou2026physinone} extend this setting with larger scene collections and denser annotations of geometry, motion, trajectories, and material parameters. Their labels are exact within the simulator, although the corresponding observations remain bounded by the chosen assets, parameter ranges, contact models, and rendering assumptions. Real-world resources offer complementary evidence. Physics 101~\cite{wu2016physics} and PhysVid~\cite{pathak2026physvid} associate controlled interaction videos with measured object properties, while ObjectFolder~\cite{gao2023object}  records visual, acoustic, and tactile responses from real and neural objects. These measurements are physically meaningful but rely on specialized acquisition protocols that limit scale and interaction diversity. Current physical supervision is consequently distributed across separate forms of evidence. Simulator datasets provide latent states without fully realistic observations, real-world datasets provide authentic dynamics with sparse property labels, and multisensory datasets rarely annotate complete state transitions. Object attributes, localized events, intermediate responses, final outcomes, and synchronized audiovisual evidence are still seldom available within the same example, leaving the physical cause of an observed transition only partially specified.

\subsection{Physics-Oriented Reinforcement Learning}

Physics-oriented reinforcement learning has emerged from the broader use of verifiable rewards for reasoning. GRPO~\cite{shao2024deepseekmathpushinglimitsmathematical} estimates relative advantages from multiple responses to the same prompt without a learned critic, while subsequent variants such as DAPO~\cite{yu2025dapo} and GSPO~\cite{zheng2025gspo} improve optimization stability through dynamic sampling, asymmetric clipping, and sequence-level objectives. This paradigm has also been extended to multimodal reasoning, where Visual-RFT~\cite{liu2025visualrftvisualreinforcementfinetuning}, VLM-R1~\cite{shen2025vlmr1}, Vision-R1~\cite{ICLR2026_680b2a81}, and R1-VL~\cite{Zhang_2025_ICCV} use automatically verifiable, rule-based, or step-wise rewards to strengthen visual reasoning without relying on a learned critic. Building on these advances, recent work has begun to incorporate physical criteria into group-relative optimization. Cosmos-Reason1~\cite{nvidia2025cosmosreason1physicalcommonsense} converts physical-common-sense and embodied-reasoning data into multiple-choice problems and optimizes answer and format rewards. HCM-GRPO~\cite{hu2026physicalhcmgrpo} mines difficult samples for physical-plausibility discrimination, reflecting evidence that group-relative optimization benefits from prompts whose sampled responses exhibit meaningful reward variation~\cite{pikus2025hard}, while Physics-R1~\cite{yang2026physicsr1} further incorporates verifiable answer and unit-consistency rewards for visual physics reasoning. Process-oriented studies supplement terminal accuracy with judgments of physical principles, units, reasoning quality, visual attention, or an explicitly constructed physical model~\cite{lilienthal2026reward,zhang2026decoupled}. 

Verifiability in these methods is often obtained by narrowing the task to answer matching, static plausibility classification, or rubric-based assessment of diagrammatic problems. Such rewards can distinguish a correct response from an incorrect one, but they provide limited evidence that the response rests on the correct material property, contact event, or state transition observed in the input. The group-relative signal is also weak when all rollouts from a prompt receive the same reward, which makes prompt selection part of the optimization problem rather than a separate data-preparation choice. Interaction-based policy learning further shows that improvements within a training environment do not necessarily yield physical knowledge that transfers to related settings~\cite{buschoff2026can}. Existing strategies have improved answer-level reliability, while perceptually grounded and temporally resolved reward signals for multimodal physical reasoning remain underdeveloped.

\section{Training Data Construction} 
\label{sec:physics_data}

\begin{wrapfigure}[15]{r}{0.38\textwidth}
    \vspace{-10pt}
    \centering
    \includegraphics[width=\linewidth]
    {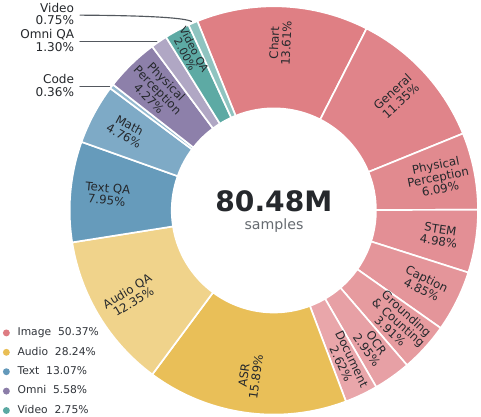}
    \caption{Training-data distribution across modalities and task categories.}
    \label{fig:training_data_distribution}
\end{wrapfigure}
Our training-data pipeline contains two components. We first construct physics-aware supervised data to provide explicit object-property and physical-event supervision, and then curate reinforcement-learning data according to the response distribution of the current policy. Figure~\ref{fig:training_data_distribution} provides an overview of the modality and task distribution of the final training corpus.

\subsection{Physics-Aware Data Construction}

Generic image and video corpora describe objects and actions but rarely provide numerical properties or cross-modal physical explanations. We therefore construct complementary static and dynamic data: the static branch performs object filtering, profiling, prototype retrieval, and property refinement to characterize material, body type, and intrinsic properties, while the dynamic branch performs event filtering, clip selection, and dense multimodal annotation to capture interactions, state changes, and acoustic evidence. Branch-specific templates then guide the same commercial large language model to convert both forms of metadata into question--answer pairs, bridging semantic perception and physical reasoning as shown in Figure~\ref{fig:datappl}.

\subsubsection{Static Physical Data Construction}
\label{sec:static_physics_data}

\begin{figure*}[t]
    \centering
    \includegraphics[width=1.0\linewidth]{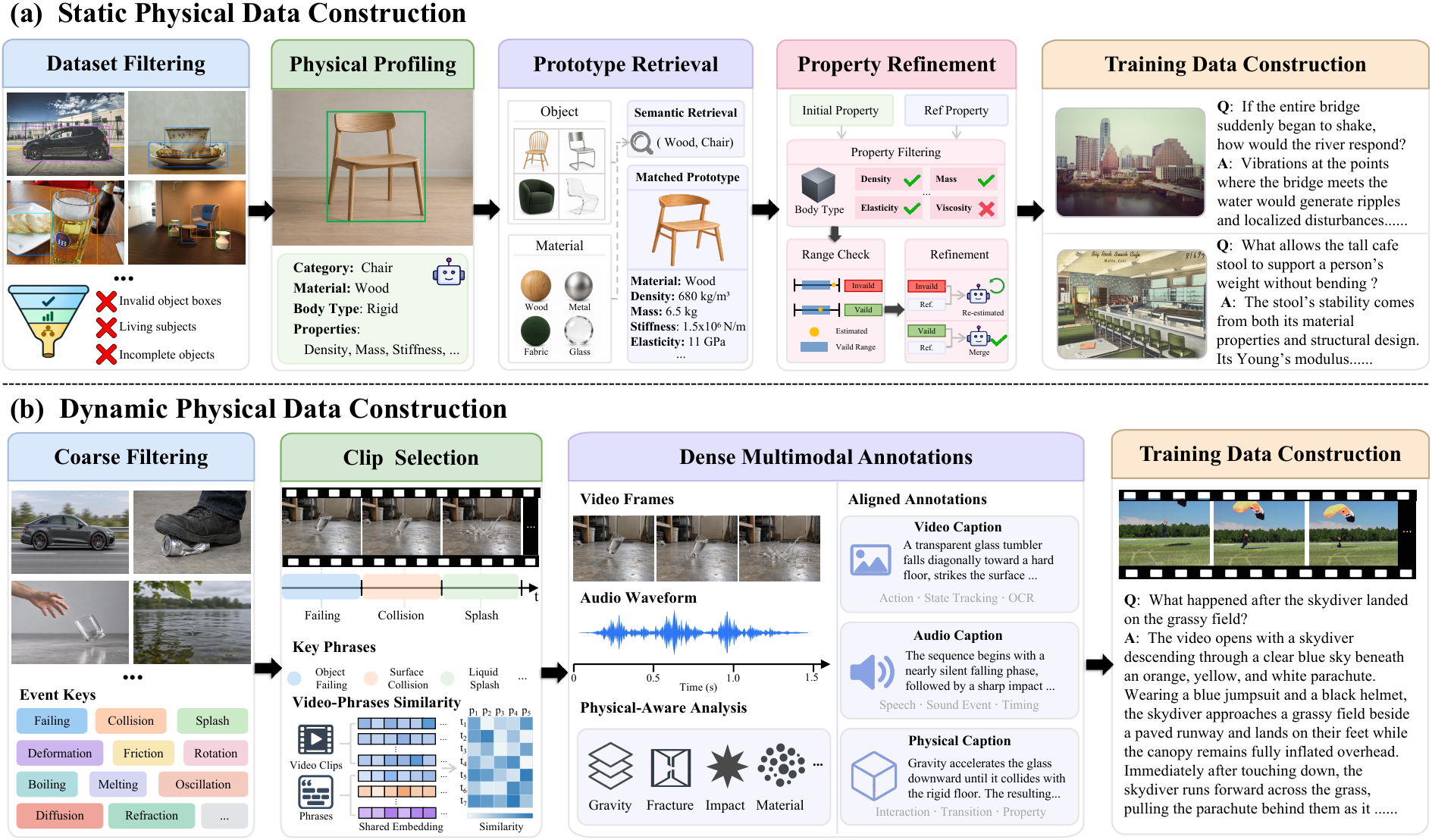}
    \caption{Physics-aware data construction. (a) The static branch filters images, identifies objects and materials, retrieves object--material prototypes, and refines physical properties through compatibility and range checks before constructing training examples. (b) The dynamic branch retrieves event-centered video clips and aligns visual descriptions, acoustic evidence, and physical analysis into temporally grounded supervision of interactions and state changes. Together, the two branches connect intrinsic object properties with their observable effects during physical events.}
    \label{fig:datappl}
\end{figure*}

\noindent\textbf{Dataset Filtering.}
As shown in the first block of Figure~\ref{fig:datappl}(a), we begin with object-level image annotations and remove invalid bounding boxes, living subjects, and severely incomplete objects. This step retains non-living objects whose visible extent is sufficient for estimating material and physical properties, while preserving the surrounding image as contextual evidence.

\noindent\textbf{Physical Profiling.}
For every retained object, its image $I_i$ and target bounding box $b_{ij}$ are provided to a commercial multimodal large model. The model returns a structured profile containing the object category $c_{ij}$, material $m_{ij}$, body type $\tau_{ij}$, and a set of applicable numerical properties:
\begin{equation}
    \mathcal{M}_{\mathrm{com}}(I_i,b_{ij})=
    \left(c_{ij},m_{ij},\tau_{ij},
    \{(k,\hat p_{ijk},u_k)\}_{k\in\mathcal{K}_{ij}}\right),
    \label{eq:physical_profile}
\end{equation}
where $\hat p_{ijk}$ and $u_k$ denote the estimated value and canonical unit of property $k$. Depending on the object and body type, $\mathcal{K}_{ij}$ may include mass, density, stiffness, friction, restitution, Young's modulus, Poisson's ratio, viscosity, surface tension, and yield stress. The values in this initial profile are produced directly by the model from the image and bounding box; the prototype bank is not used at this stage.

\noindent\textbf{Prototype Retrieval.}
We build a reference bank containing  1,209 object--material prototypes. Each prototype stores normalized category and material keywords, body type, and real-value references or valid intervals compiled from material handbooks and representative product specifications. Given the predicted category and material, we perform keyword matching over the two indices and retrieve
\begin{equation}
    r_{ij}^{\star}=\arg\max_{r\in\mathcal{R}}
    \left[\lambda_c J(\mathcal{W}_{c_{ij}},\mathcal{W}_{c_r})+
    \lambda_m J(\mathcal{W}_{m_{ij}},\mathcal{W}_{m_r})\right],
    \label{eq:prototype_retrieval}
\end{equation}
where $\mathcal{W}$ denotes a normalized keyword set and $J(\cdot,\cdot)$ is Jaccard overlap. Category keywords identify a functionally similar object prototype, while material keywords provide the corresponding material-level reference. The matched entry supplies a reference value and a valid interval $[l_{rk},u_{rk}]$ for each applicable property.

\noindent\textbf{Property Refinement.}
We first filter out attributes that are irrelevant to the inferred body type, such as viscosity for a rigid chair. Each remaining estimate is then compared with the interval of the matched prototype. An out-of-range value is returned to the same commercial model together with the reference property and interval for one re-estimation:
\begin{equation}
    \begin{aligned}
    \tilde p_{ijk}&=\operatorname{Refine}_{\mathcal{M}_{\mathrm{com}}}
    (I_i,b_{ij},\hat p_{ijk},[l_{rk},u_{rk}]),\\
    p_{ijk}^{\star}&=
    \begin{cases}
        \hat p_{ijk}, & \hat p_{ijk}\in[l_{rk},u_{rk}],\\
        \tilde p_{ijk}, & \tilde p_{ijk}\in[l_{rk},u_{rk}],\\
        \varnothing, & \text{otherwise}.
    \end{cases}
    \end{aligned}
    \label{eq:property_refinement}
\end{equation}
Only one refinement pass is performed. If a required value remains outside the valid interval after receiving the reference, the entire object sample is discarded; otherwise, the accepted and re-estimated values are merged into its final physical profile.

\noindent\textbf{Training Data Construction.}
The refined object metadata are organized with static-data templates and provided to the commercial large language model, which summarizes them into question--answer pairs for physical-property estimation and property-grounded reasoning. The corresponding image is attached to each generated instruction as its visual input.

\subsubsection{Dynamic Physical Data Construction}
\label{sec:dynamic_physics_data}

\noindent\textbf{Coarse Filtering.}
As shown in the first block of Figure~\ref{fig:datappl}(b), we construct a hierarchical physical-event library comprising six coarse categories and 26 fine-grained event types. Each event type is associated with a set of retrieval keywords, which are matched against the available video titles, descriptions, and tags to remove videos without an explicit physical event. We further expand the retained keywords into short phrases that combine an object, an interaction, and an observable result.

\noindent\textbf{Clip Selection.}
Long videos often contain only a short interval relevant to the event key. We divide each candidate video into overlapping temporal clips and use Qwen3-VL-Embedding-8B~\cite{li2026qwen3vlembedding} to encode both the clips and the expanded phrases in a shared embedding space. For clip $v_i$ and phrase $q_j$, their matching score is
\begin{equation}
    s_{ij}=\frac{E_v(v_i)^{\top}E_t(q_j)}
    {\lVert E_v(v_i)\rVert_2\lVert E_t(q_j)\rVert_2},
    \qquad
    v_j^{\star}=\arg\max_{v_i}s_{ij},
    \label{eq:clip_selection}
\end{equation}
where $E_v$ and $E_t$ are the video and text encoders. The top-matched neighboring clips are merged to preserve the pre-event state, interaction, and visible consequence. 

\noindent\textbf{Dense Multimodal Annotations.}
The selected clip, sampled video frames, and synchronized audio waveform are jointly provided to the commercial multimodal large model. The model first produces separate descriptions of the visual and acoustic streams. The visual description records actions, object states, trajectories, and visible text, while the audio description identifies speech, sound events, and their timestamps. It then combines the two streams with physical priors to describe the underlying interaction, state change, and relevant physical properties. We keep the three descriptions separate so that the physical interpretation can be checked against the original visual and acoustic evidence.

For an event $e$ with visual onset $t_e^{\mathrm{v}}$ and acoustic onset $t_e^{\mathrm{a}}$, we associate the two observations only when their temporal discrepancy satisfies
\begin{equation}
    \Delta t_e=\left|t_e^{\mathrm{v}}-t_e^{\mathrm{a}}\right|\leq\delta_e,
    \label{eq:dense_alignment}
\end{equation}
where $\delta_e$ is an event-dependent tolerance. The aligned record therefore preserves the video description, audio description, physical interpretation, and their time anchors instead of collapsing them into a generic video caption.

\noindent\textbf{Training Data Construction.}
The aligned video, audio, and physical annotations are organized with dynamic-data templates and provided to the commercial large language model, which summarizes them into question--answer pairs for event description, temporal reasoning, and physical explanation. The source video and audio are attached to the generated instruction as its multimodal input.

\begin{figure*}[htbp]
    \centering
    \includegraphics[width=1.0\linewidth]{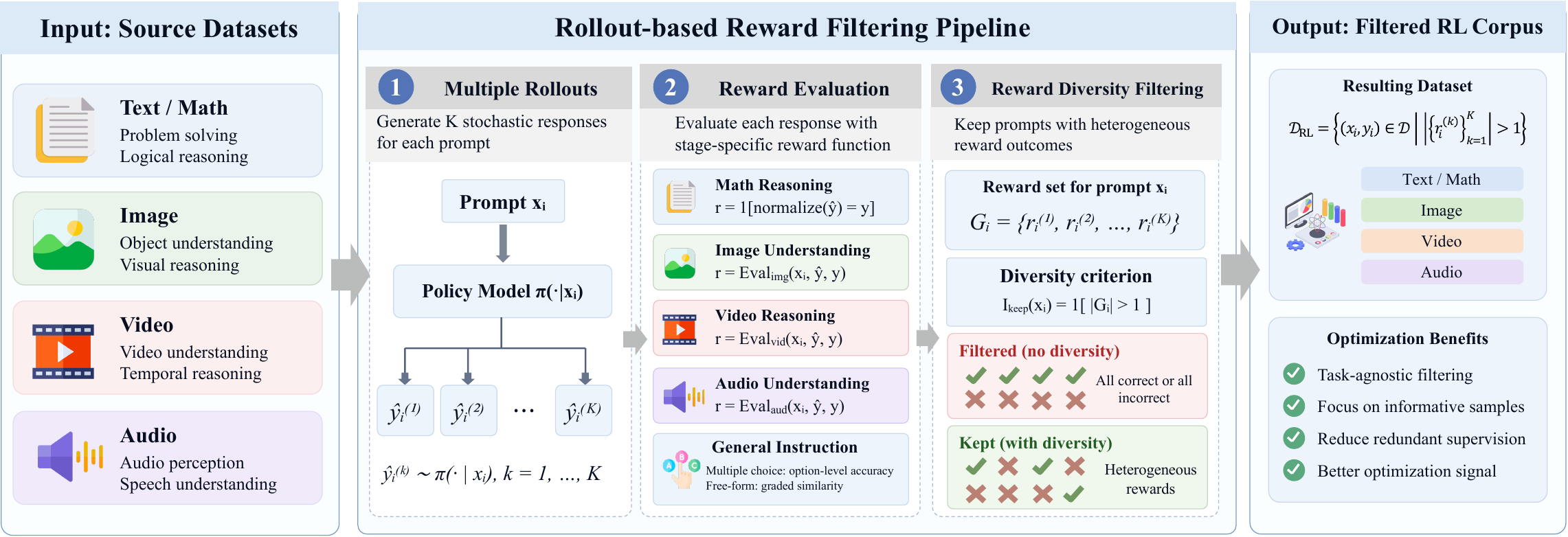}
    \caption{Rollout-based reward-diversity filtering for RL data curation. For each prompt, a fixed policy generates $K$ stochastic responses ($K=4$ by default), which are scored by task-specific reward evaluators. A prompt is retained only if its rollout group contains more than one distinct reward value. Uniform-reward groups are discarded; for binary rewards, these correspond to all-correct or all-incorrect responses. The retained corpus provides within-group reward variation for subsequent GRPO training.}
    \label{fig:rl_data}
\end{figure*}

\subsection{Policy-Aware RL Data Curation}
\label{sec:rl_data_curation}

Unlike supervised examples, the utility of an RL prompt depends on both the current policy and the optimization rule. GRPO ranks responses sampled for the same prompt through group-normalized rewards (Eq.~\eqref{eq:grpo_advantage}). If every response in a sampled group receives the same reward, all normalized advantages are zero, so that group provides no reward-based preference for the corresponding update. We therefore curate the RL corpus according to the response distribution of a fixed rollout policy rather than treating every candidate prompt as equally informative. Figure~\ref{fig:rl_data} summarizes this procedure across candidate pools spanning text, image, video, and audio inputs.

\noindent\textbf{Group-Wise Rollout Evaluation.}
Let \(\mathcal{D}=\{(x_i,y_i)\}_{i=1}^{N}\) denote a candidate pool, where \(x_i\) is a prompt and \(y_i\) is its reference answer. For each \(x_i\), a fixed policy \(\pi_{\mathrm{roll}}\) generates \(K\) stochastic responses under the same decoding configuration, with \(K=4\) by default. Each response is scored independently by the evaluator \(\mathcal{R}_{\tau_i}\) associated with its task type \(\tau_i\):
\begin{equation}
\begin{aligned}
\hat y_i^{(k)} &\sim \pi_{\mathrm{roll}}(\cdot\mid x_i),\\
r_i^{(k)} &= \mathcal{R}_{\tau_i}
\!\left(x_i,\hat y_i^{(k)},y_i\right),
\qquad k=1,\ldots,K.
\end{aligned}
\label{eq:curation_rollout}
\end{equation}
Keeping the rollout policy and decoding configuration fixed ensures that the reward diversity within each rollout group is induced by the policy's stochastic response behavior under a consistent sampling process.

\noindent\textbf{Reward-Diversity Selection.}
To identify prompts that provide informative group-relative learning signals, we retain a prompt only if its sampled responses receive at least two distinct reward values. With \(\mathcal{S}_i=\{r_i^{(k)}\}_{k=1}^{K}\), the selection rule and resulting corpus are
\begin{equation}
\begin{aligned}
q_i &= \mathbb{I}\!\left[|\mathcal{S}_i|>1\right],\\
\mathcal{D}_{\mathrm{RL}} &=
\left\{(x_i,y_i)\in\mathcal{D}\mid q_i=1\right\}.
\end{aligned}
\label{eq:reward_diversity_filter}
\end{equation}
For a binary correctness reward, this retains groups containing both correct and incorrect responses and removes both all-correct and all-incorrect groups. For a graded or composite reward, the same rule retains any group with more than one attained score. The criterion thus mirrors the requirement of group-relative optimization: a retained prompt exhibits an observable reward ordering under the rollout policy.

\noindent\textbf{Task-Specific Reward Instantiation.}
The selection rule is shared across data sources, while the evaluator follows the intended training objective. Mathematical and symbolic answers are checked for equivalence, with normalized string matching as a fallback. Multiple-choice image--text, video, and mixed-modal tasks use exact option matching, whereas free-form answers use their task-specific normalized or graded matching rules; malformed or degenerate generations are rejected.

Candidate data for progressive multimodal RL are scored by final-answer correctness, while the physical-perception data used in the fine-grained RL phase are scored with the answer, intermediate-perception, and format terms defined in Eq.~\eqref{eq:fine_grained_reward}. Thus, task-specific evaluators determine response quality, and the shared diversity rule determines whether that quality difference is observable within a rollout group.

\section{Method}
\subsection{Model Architecture}
\label{sec:model}

\begin{figure*}[t]
    \centering
    \includegraphics[width=1.0\linewidth]{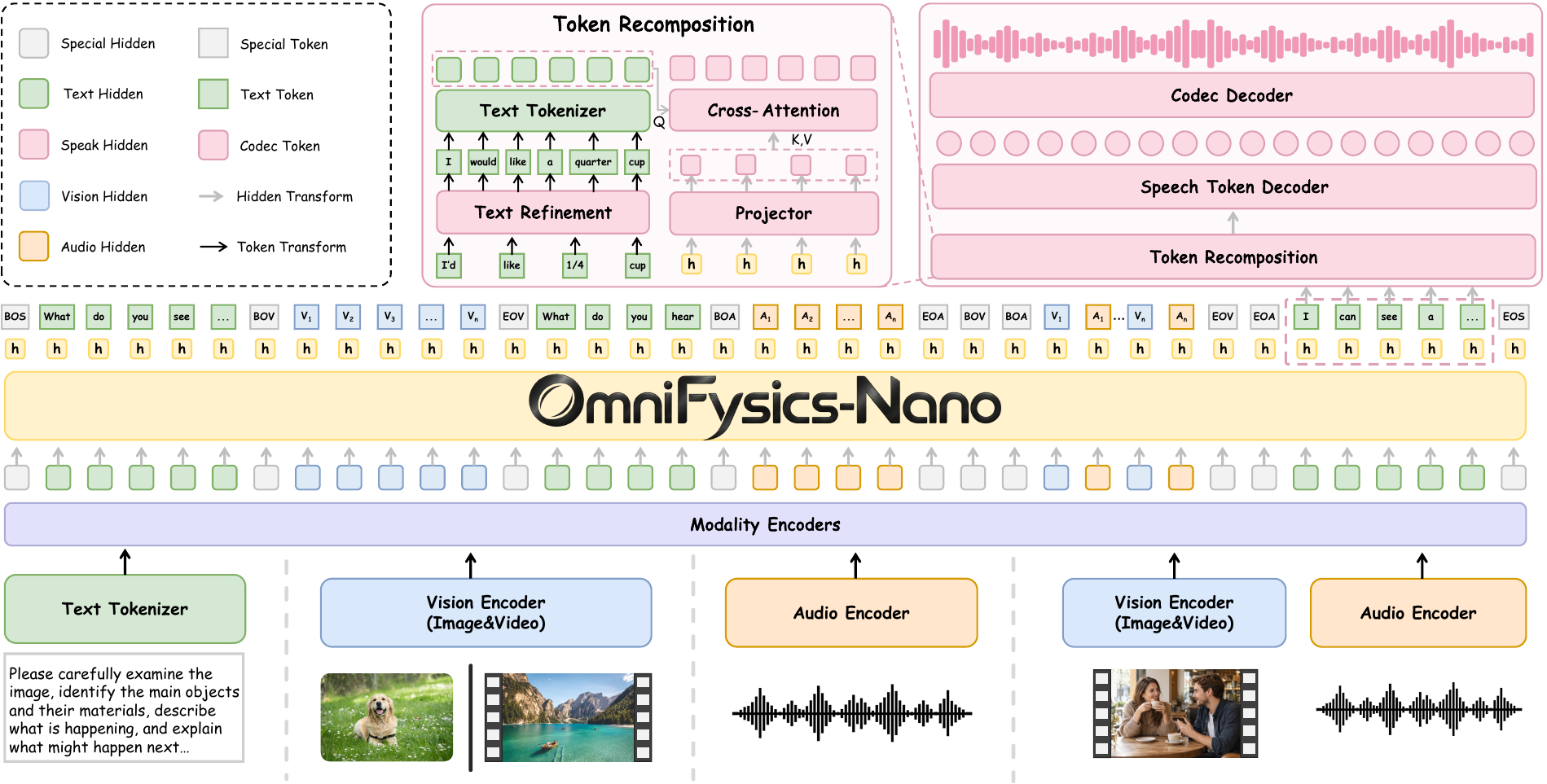}
    \caption{Architecture of OmniFysics-Nano-V2. Text, visual, and acoustic features are packed into a unified sequence for a shared causal backbone; temporally corresponding visual and acoustic groups are interleaved for synchronized video--audio input. For speech output, the generated answer is normalized and retokenized with the speech tokenizer. The resulting text embeddings query the projected answer states through cross-attention, producing a semantic conditioning sequence for speech-token generation and waveform reconstruction.}
    \label{fig:model_architecture}
\end{figure*}

As illustrated in Figure~\ref{fig:model_architecture}, OmniFysics-Nano-V2 follows a unified perception--reasoning--expression design. Modality interfaces first map heterogeneous observations into a common token space, where a shared causal transformer performs cross-modal reasoning. The generated answer states are then reused to condition speech synthesis through cross-attention. This separation keeps semantic reasoning in a single pathway while allowing the same response to be expressed as either text or speech.

\subsubsection{Unified Omni-Modal Sequence}
Let $x_{\mathrm{txt}}$, $x_{\mathrm{vis}}$, and $x_{\mathrm{aud}}$ denote text, image or video, and audio inputs. Their representations are
\begin{equation}
\begin{aligned}
X_{\mathrm{txt}}&=E_{\mathrm{txt}}(x_{\mathrm{txt}}),\\
X_{\mathrm{vis}}&=M_{\mathrm{vis}}\!\left(E_{\mathrm{vis}}(x_{\mathrm{vis}})\right),\\
X_{\mathrm{aud}}&=P_{\mathrm{aud}}\!\left(E_{\mathrm{aud}}(x_{\mathrm{aud}})\right),
\end{aligned}
\label{eq:modality_interfaces}
\end{equation}
where $E_{\mathrm{txt}}$, $E_{\mathrm{vis}}$, and $E_{\mathrm{aud}}$ are the text, visual, and acoustic encoders, respectively; $M_{\mathrm{vis}}$ is the visual merger; and $P_{\mathrm{aud}}$ projects acoustic features to the shared hidden dimension. Images and videos use the same visual interface, preserving spatial layout and frame order, while audio and speech share the acoustic interface. Modality-specific boundary embeddings explicitly mark the extent of every non-text segment.

The modality-specific representations are inserted at their corresponding positions in the conversational sequence, delimited by modality boundary tokens, and jointly processed by the shared causal transformer $F_{\theta}$. Consequently, each response token can condition on textual, visual, and acoustic evidence within a single autoregressive context.

For synchronized video--audio input, the visual and acoustic streams are divided into temporally corresponding feature groups and interleaved as
\begin{equation}
[b_{\mathrm{vis}},b_{\mathrm{aud}},X_{\mathrm{vis}}^{(1)},X_{\mathrm{aud}}^{(1)},\ldots,
X_{\mathrm{vis}}^{(N)},X_{\mathrm{aud}}^{(N)},e_{\mathrm{vis}},e_{\mathrm{aud}}].
\label{eq:av_interleave}
\end{equation}
Each group may contain multiple tokens; the superscript indexes a shared temporal interval rather than a one-to-one token correspondence. This ordering places temporally related visual and acoustic evidence in proximity while preserving the internal structure of each modality, making event-level correspondences directly accessible to the causal backbone.

\subsubsection{Semantic Alignment for Speech Generation}
\label{sec:speech_output}
Let $a=(a_1,\ldots,a_{T_a})$ be the textual answer generated by the reasoning backbone and $H_a\in\mathbb{R}^{T_a\times d}$ its final-layer hidden states. Rather than conditioning speech generation on the rendered answer alone, we retain these answer states as an additional semantic representation. A trainable MLP $P_{\phi}$ maps them to the speech-conditioning space, followed by feature normalization:
\begin{equation}
Z_a=\operatorname{Norm}\!\left(P_{\phi}(H_a)\right).
\label{eq:answer_projection}
\end{equation}
In parallel, detokenization recovers the response string from the generated text tokens. Text normalization then converts the raw response into a canonical, human-readable and speakable form by removing non-spoken formatting and rewriting symbols or numerals when necessary. The normalized text is subsequently processed by the speech-side tokenizer:
\begin{equation}
C=\tau_{\mathrm{sp}}\!\left(\mathcal{N}\!\left(\operatorname{Detok}_{\mathrm{txt}}(a)\right)\right),
\qquad E_C=\operatorname{Emb}_{\mathrm{sp}}(C),
\label{eq:speech_retokenization}
\end{equation}
where $C=(c_1,\ldots,c_{T_c})$. Because the language and speech tokenizers segment the same answer differently, $T_a$ and $T_c$ need not match and their representations cannot be aligned position by position. We therefore apply cross-attention between the speech-side text embeddings and the projected answer states:
\begin{equation}
U=E_C+\operatorname{CrossAttn}(E_C,Z_a).
\label{eq:answer_cross_attention}
\end{equation}
The residual connection preserves the speech-side textual representation, while cross-attention adds context from the complete sequence of answer states. The resulting sequence $U$ is combined with the speech start token, control-text embeddings, and task identifier to form the conditioning prefix
\begin{equation}
\Pi=[e_{\mathrm{sos}},e_{\mathrm{ctrl}},U,e_{\mathrm{task}}].
\label{eq:speech_prefix}
\end{equation}
Conditioned on $\Pi$, the speech-token decoder $D_{\psi}$ predicts a discrete speech-token sequence autoregressively until the end-of-sequence token is reached. After token generation is complete, the codec decoder $G_{\xi}$ converts the discrete sequence into the output waveform:
\begin{equation}
\begin{aligned}
s_t&\sim p_{\psi}(\cdot\mid s_{<t},\Pi),\quad t=1,\ldots,T_s,\\
\hat{w}&=G_{\xi}(s_{1:T_s}).
\end{aligned}
\label{eq:speech_generation}
\end{equation}
This decomposition assigns semantic transfer to cross-attention, sequence modeling to the speech-token decoder, and waveform reconstruction to the codec decoder. It also allows speech generation to reuse the states that produced the written answer rather than recovering its semantics from text alone.

\subsubsection{Training Objective}
The model is trained with the causal language-modeling objective, with the speech term activated only during the final output-alignment stage:
\begin{equation}
\mathcal{L}_{\mathrm{SFT}}=\mathcal{L}_{\mathrm{text}}+\lambda_{\mathrm{sp}}\mathcal{L}_{\mathrm{speech}},
\label{eq:sft_loss}
\end{equation}
where $\lambda_{\mathrm{sp}}=0$ for multimodal supervised fine-tuning and is nonzero for speech alignment. This objective keeps the main model unchanged while the output branch learns its own interface.

\begin{figure*}[t]
    \centering
    \includegraphics[width=1.0\linewidth]{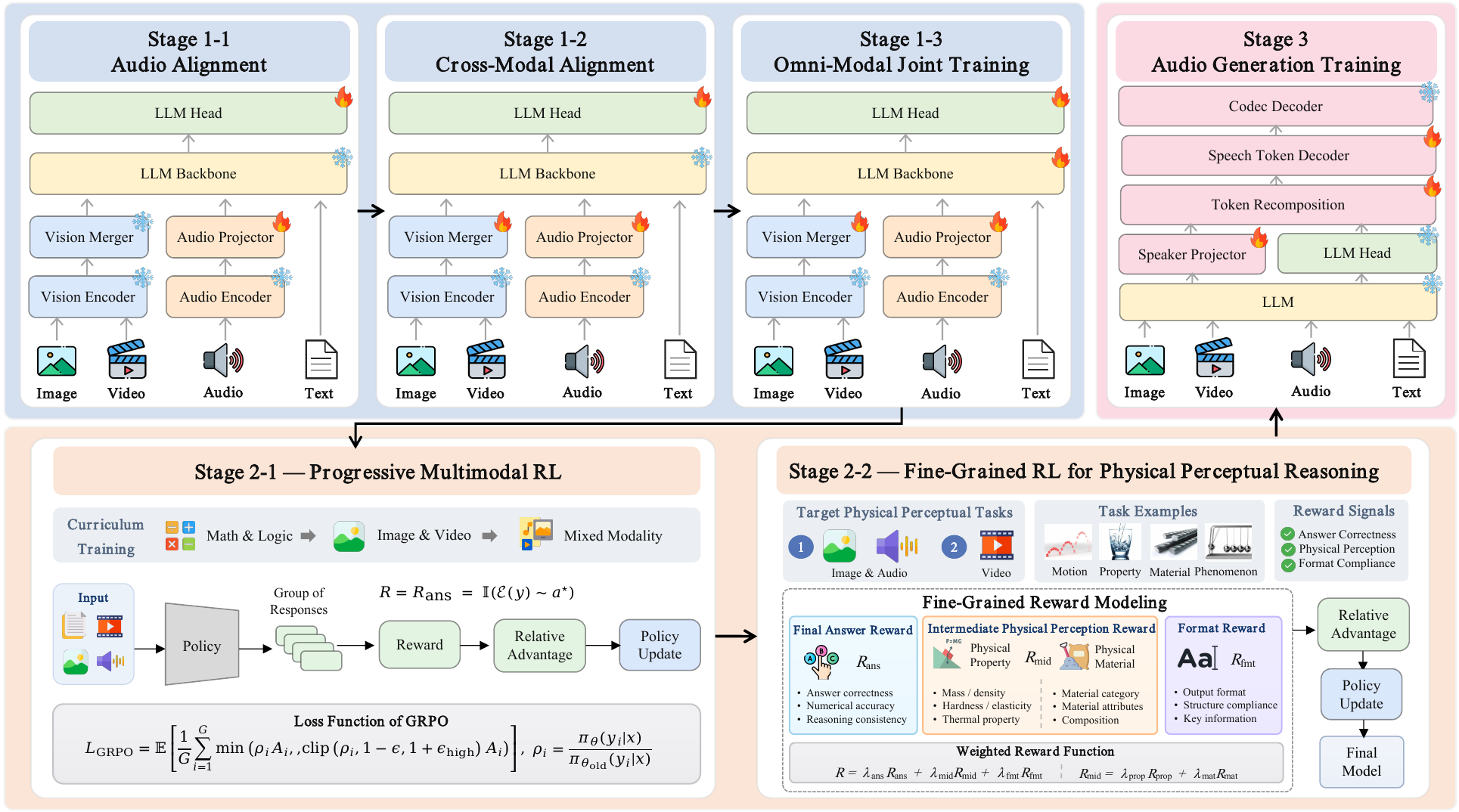}
    \caption{Training curriculum of OmniFysics-Nano-V2. Stage 1 comprises Audio Alignment (Stage 1-1), Cross-Modal Alignment (Stage 1-2), and Omni-Modal Joint Training (Stage 1-3). Stage 2 first performs Progressive Multimodal RL with answer-level rewards (Stage 2-1), then applies Fine-Grained RL for Physical Perceptual Reasoning with final-answer, intermediate-perception, and format rewards (Stage 2-2). Stage 3 performs Audio Generation Training while keeping the multimodal reasoner and codec decoder frozen. Flame and snowflake symbols denote trainable and frozen components, respectively.}
    \label{fig:training}
\end{figure*}

\subsection{Training Strategy}
Our training strategy combines supervised fine-tuning (SFT) and reinforcement learning (RL) within a staged curriculum. We first define the optimization objectives shared across stages and then describe how they are assigned to different model components and data mixtures.

\subsubsection{Optimization Objectives}
\noindent\textbf{Supervised Fine-Tuning.} Given a multimodal context $x$ and target response $a=(a_1,\ldots,a_{T_a})$, SFT minimizes the standard autoregressive next-token loss
\begin{equation}
\mathcal{L}_{\mathrm{text}}(\theta)=-\sum_{t=1}^{T_a}
\log p_{\theta}(a_t\mid a_{<t},x).
\label{eq:text_loss}
\end{equation}
This objective provides token-level supervision for aligning modality interfaces and learning joint multimodal responses in Stage 1.

\noindent\textbf{Group Relative Policy Optimization.} For each input $x$, the behavior policy $\pi_{\theta_{\mathrm{old}}}$ samples a group of $G$ responses $\mathcal{Y}(x)=\{y_i\}_{i=1}^{G}$. After assigning each response a stage-specific reward $r_i=R(x,y_i)$, we normalize rewards within the group to obtain
\begin{equation}
    A_i=\frac{r_i-\mu_{\mathcal{G}}}{\sigma_{\mathcal{G}}+\delta},
    \qquad
    \mu_{\mathcal{G}}=\frac{1}{G}\sum_{j=1}^{G}r_j,
    \label{eq:grpo_advantage}
\end{equation}
where $\sigma_{\mathcal{G}}$ is the group-wise standard deviation and $\delta$ prevents division by zero. We maximize the clipped sequence-level objective
\begin{equation}
\begin{aligned}
\mathcal{J}_{\mathrm{GRPO}}=\mathbb{E}\Bigg[\frac{1}{G}\sum_{i=1}^{G}
\min\Big(&\rho_i A_i,\operatorname{clip}(\rho_i,1-\epsilon,1+\epsilon_{\mathrm{high}})A_i\Big)\Bigg],\\
\rho_i={}&\frac{\pi_{\theta}(y_i\mid x)}
{\pi_{\theta_{\mathrm{old}}}(y_i\mid x)},
\end{aligned}
\label{eq:grpo_objective}
\end{equation}
Here $\rho_i$ is computed over the complete response rather than independently for each token. Group normalization makes the update depend on the relative quality of responses sampled for the same input, consistent with the reward-diversity filtering used to construct the RL corpus.

\noindent\textbf{Speech Generation.} For target speech tokens $s_{1:T_s}$, let $y_{1:T_s}=s_{1:T_s}$ and append $y_{T_s+1}=s_{\mathrm{eos}}$. Conditioned on the prefix $\Pi$ from Eq.~\eqref{eq:speech_prefix}, the answer projector and cross-attention parameters $\omega$ are jointly optimized with the speech-token decoder parameters $\psi$ through
\begin{equation}
\mathcal{L}_{\mathrm{speech}}(\psi,\omega)=-\sum_{t=1}^{T_s+1}
\log p_{\psi,\omega}(y_t\mid y_{<t},\Pi).
\label{eq:speech_loss}
\end{equation}
This objective jointly optimizes the answer-conditioning module $\omega$ and the speech-token decoder $\psi$.

\subsubsection{Stage-Wise Training}
Figure~\ref{fig:training} summarizes our three-stage curriculum. Stage 1 establishes multimodal input alignment and joint understanding through three supervised phases. Stage 2 optimizes the resulting policy through progressive and fine-grained reinforcement learning. Stage 3 adds speech generation after the text policy has been optimized. Every phase is initialized from the preceding checkpoint. This ordering separates interface alignment, policy optimization, and output adaptation, reducing interference between objectives that act on different parts of the model.

\noindent$\bullet$ \textbf{Stage 1: Supervised Multimodal Training.} All three phases optimize the SFT objective in Eq.~\eqref{eq:text_loss}, while progressively expanding the trainable components and supervision mixture.

\noindent\textbf{Stage 1-1: Audio Alignment.} We first use ASR supervision to align the acoustic representation with the shared language space. The acoustic and visual encoders, visual merger, and causal backbone remain frozen; only the audio projector and newly introduced audio-boundary embeddings are optimized. Constraining the update to the audio interface prevents the initial alignment objective from overwriting pretrained linguistic and visual representations.

\noindent\textbf{Stage 1-2: Cross-Modal Alignment.} Starting from the audio-aligned checkpoint, we introduce image and video supervision and optimize the visual mergers together with the audio projector and modality-boundary embeddings. Captioning, OCR, ASR, and video-caption tasks provide direct alignment signals, while the modality encoders and causal backbone remain fixed. This phase brings visual and acoustic features into the language model's representational space before joint reasoning is learned.

\noindent\textbf{Stage 1-3: Omni-Modal Joint Training.} After the modality interfaces have been aligned, we unfreeze the causal backbone and optimize it jointly with the visual mergers, audio projector, and boundary embeddings; the modality encoders remain frozen. The training mixture covers image, video, audio, speech, text, embodied, and physics-oriented instructions. At this point, the objective moves beyond interface alignment: the shared transformer learns to combine appearance, temporal change, acoustic evidence, and language context within a single autoregressive response.

\noindent$\bullet$ \textbf{Stage 2: Reinforcement Learning.} Supervised likelihood training encourages imitation but does not directly optimize verifiable reasoning outcomes. We therefore apply two consecutive GRPO phases. They share the same policy-update rule but differ in curriculum and reward granularity: Stage 2-1 uses answer-level correctness across increasingly diverse modalities, whereas Stage 2-2 adds explicit feedback for intermediate physical perception and response format.

\noindent\textbf{Stage 2-1: Progressive Multimodal RL.} This phase uses a binary final-answer reward. Let $\mathcal{E}(y)$ extract the predicted answer from response $y$, let $a^\star$ be the reference answer, and let $\mathcal{M}_{\tau(x)}$ denote the verifier associated with the task type $\tau(x)$. We define
\begin{equation}
R_{\mathrm{ans}}(x,y)=
\mathbb{I}\!\left[\mathcal{M}_{\tau(x)}\!\left(\mathcal{E}(y),a^\star\right)=1\right].
\label{eq:answer_reward}
\end{equation}
For mathematical and symbolic tasks, $\mathcal{M}_{\tau}$ checks symbolic equivalence and falls back to normalized string matching when parsing fails. Multiple-choice tasks require an exact option match, while OCR and free-form multimodal tasks compare normalized answers using their task-specific evaluators. The verifier therefore changes with the answer format, but the optimization signal remains final-answer correctness.

Training proceeds from mathematical and logical reasoning to image and video reasoning, and finally to mixed-modality problems involving visual and acoustic evidence. The first phase establishes reasoning and answer verification in a controlled text setting; subsequent phases introduce spatial, temporal, and cross-modal dependencies. Throughout this curriculum, the GRPO update and answer-level reward principle remain fixed, while the input complexity progressively increases.

\noindent\textbf{Stage 2-2: Fine-Grained RL for Physical Perceptual Reasoning.} An answer-only reward cannot distinguish a response that identifies the relevant physical evidence but makes a downstream error from one that never extracts the required evidence. We therefore introduce denser supervision for image--audio and video tasks involving motion, physical properties, materials, and physical phenomena. The reward combines final-answer correctness, intermediate physical perception, and format compliance:
\begin{equation}
\begin{aligned}
R={}&\lambda_{\mathrm{ans}}R_{\mathrm{ans}}
+\lambda_{\mathrm{mid}}R_{\mathrm{mid}}
+\lambda_{\mathrm{fmt}}R_{\mathrm{fmt}},\\
R_{\mathrm{mid}}={}&\lambda_{\mathrm{prop}}R_{\mathrm{prop}}
+\lambda_{\mathrm{mat}}R_{\mathrm{mat}},
\end{aligned}
\label{eq:fine_grained_reward}
\end{equation}
where all weights are nonnegative, $\lambda_{\mathrm{ans}}+\lambda_{\mathrm{mid}}+\lambda_{\mathrm{fmt}}=1$, and $\lambda_{\mathrm{prop}}+\lambda_{\mathrm{mat}}=1$. $R_{\mathrm{ans}}$ evaluates the final decision using the task-specific verifier. $R_{\mathrm{prop}}$ evaluates the requested physical property or attribute value, or agreement with an annotated numerical interval for quantitative predictions, while $R_{\mathrm{mat}}$ evaluates material recognition and associated attributes. $R_{\mathrm{fmt}}$ checks structural compliance and the presence of required fields; it does not substitute for semantic correctness.

For image--audio tasks, the intermediate reward evaluates whether the response identifies both the queried property and the material evidence needed to support it. For video tasks, it evaluates physical attributes inferred from temporal observations. These intermediate terms expose which perceptual step failed and provide a more localized learning signal than final-answer correctness alone.

\noindent$\bullet$ \textbf{Stage 3: Audio Generation Training.} After reinforcement learning, we freeze the modality encoders, shared causal backbone, and codec decoder so that speech adaptation cannot alter the learned multimodal policy. Only the answer projector and cross-attention parameters $\omega$ and the speech-token decoder parameters $\psi$ are jointly updated using the speech-generation objective in Eq.~\eqref{eq:speech_loss}. Because this objective is applied only after text-policy optimization, it adapts the speech-output path while preserving both the learned multimodal reasoner and waveform decoder.

\subsubsection{Joint Optimization with Differentiable Physics Engine}
We further couple OmniFysics-Nano-V2 with our in-house differentiable physics engine to calibrate the object-level physical properties inferred from visual observations. 
The model first estimates mass, density, static and kinetic friction coefficients, and restitution for each visible object. 
Together with the perceived scene geometry and initial object states, these predictions are used to parameterize the corresponding instances in the physics engine. 
The engine then performs forward simulation under the same initial conditions and action sequence as the reference interaction, producing the simulated evolution of object positions, motion trajectories, and velocities.

The simulated dynamics are compared with the reference observations in terms of terminal position, complete trajectory, and velocity. 
Because the physical engine is differentiable, the resulting discrepancy can be propagated backward through the sequence of physical transitions to the predicted physical parameters. 
These parameters are iteratively corrected while remaining within physically valid ranges, after which the updated simulation is evaluated again. 
This recurring process forms a perception--simulation--calibration loop that uses observed motion to refine the model's initial estimates, converting visual physical priors into physically consistent, simulation-ready parameters for downstream policy optimization.

\section{Experiments and Results}
\label{sec:Experiments}

\begin{table*}[t]
    \centering
    \small
    \caption{General multimodal understanding on MMBench-V1.1, MMStar, MMMU, HallusionBench, and AI2D. Models are grouped as vision-language (VL) or omni-modal (Omni); Size is the reported parameter count in billions (B). Higher scores are better. Bold and underlined values denote the best and second-best results, respectively, in each column; the shaded row is our model.}
    \setlength{\tabcolsep}{4pt}
    \begin{tabular}{l c c c c c c c}
        \toprule
        \textbf{Model} & \textbf{Size} & \textbf{Modality} &
        \textbf{MMBench-V1.1} & \textbf{MMStar} & \textbf{MMMU} &
        \textbf{HallusionBench} & \textbf{AI2D} \\
        \midrule

        Ovis2.5~\cite{lu2025ovis25} & 2B & VL &
        79.20 & 67.70 & 58.70 & 58.50 & 85.00 \\

        SAIL-VL2~\cite{yin2025sailvl2} & 2B & VL &
        80.10 & 64.00 & 49.30 & 51.10 & 83.10 \\

        Ovis-U1~\cite{wang2025ovisu1} & 3B & VL &
        77.90 & 61.30 & 50.60 & 55.80 & 85.60 \\

        Qwen3.5-4B~\cite{qwen2026qwen35} & 4B & VL &
        84.30 & \underline{69.70} & \textbf{77.60} &
        \underline{59.90} & 84.60 \\

        Qwen3-VL-4B-Instruct~\cite{bai2025qwen3vl} & 4B & VL &
        82.20 & 63.70 & 56.40 & 55.20 & 78.20 \\

        \midrule

        Qwen2.5-Omni-3B~\cite{xu2025qwen25omni} & 3B & Omni &
        77.80 & 55.70 & 53.10 & 40.21 & 79.50 \\

        Qwen2.5-Omni-7B~\cite{xu2025qwen25omni} & 7B & Omni &
        81.80 & 64.00 & 59.20 & 44.90 & 83.20 \\

        OmniVinci~\cite{ye2025omnivinci} & 7B & Omni &
        \textbf{88.50} & 64.50 & 49.70 & 33.00 & \textbf{91.50} \\

        \rowcolor{gray!15}
        \textbf{OmniFysics-Nano-V2} & 4B & Omni &
        \underline{85.53} & \textbf{75.27} & \underline{62.11} &
        \textbf{62.49} & \underline{88.60} \\

        \bottomrule
    \end{tabular}
    \label{tab:main_results}
\end{table*}


\begin{figure*}[t]
    \centering¡
    \includegraphics[width=1.0\linewidth]{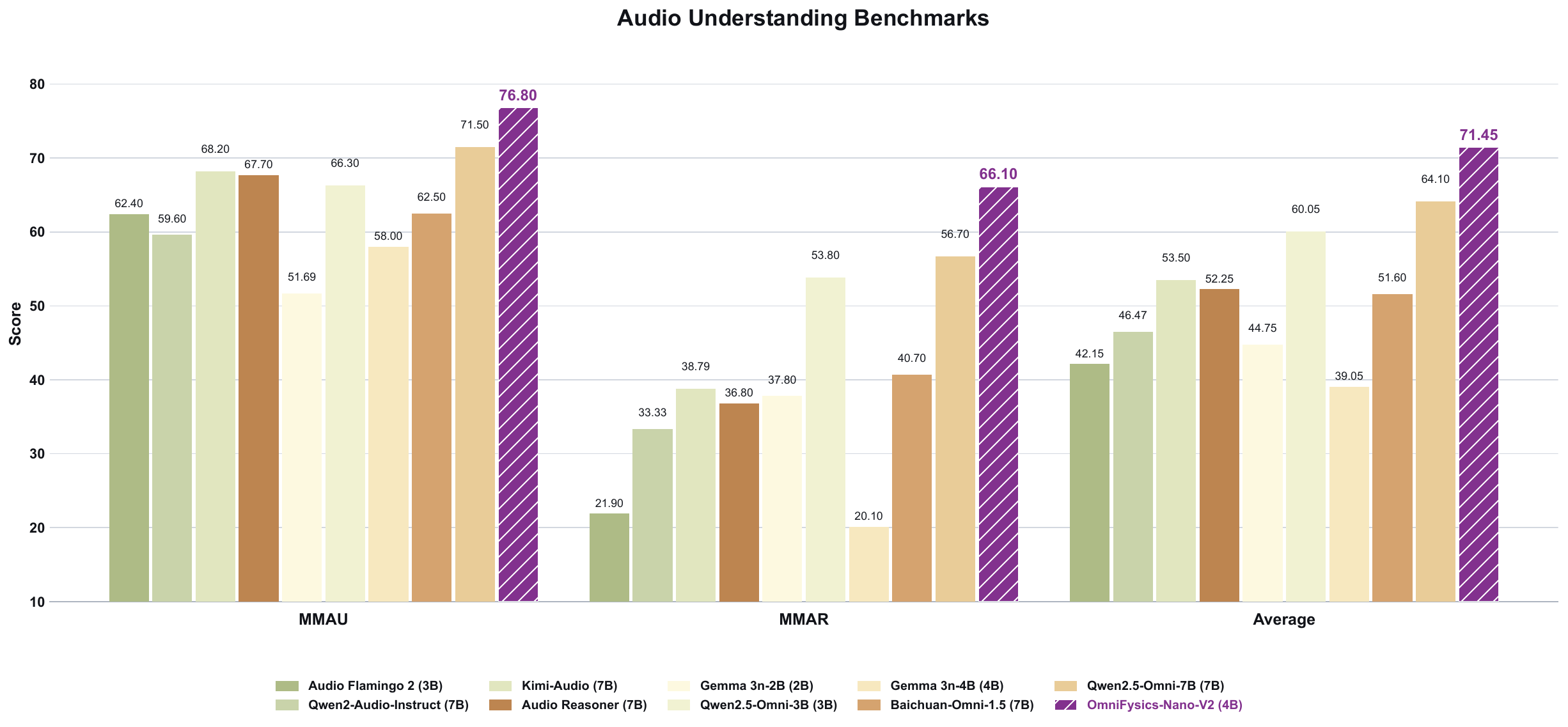}
    \caption{Comparison of Audio Understanding Performance on MMAU and MMAR. Purple hatched bars denote our model.}
    \label{fig:audio_understanding_bar}
\end{figure*}

\begin{table*}[t]
    \centering
    \small
    \caption{Omni-modal and video understanding on OmniBench, WorldSense, Daily-Omni, FysicsWorld, and Video-MME. Size is the reported parameter count in billions (B). Higher scores are better. Bold and underlined values denote the best and second-best results, respectively, in each column; the shaded row is our model.}
    \label{tab:omni_results}
    \setlength{\tabcolsep}{3.5pt}
    \begin{tabular}{l c c c c c c c}
        \toprule
        \textbf{Model} & \textbf{Size} &
        \textbf{OmniBench} & \textbf{WorldSense} & \textbf{Daily-Omni} &
        \textbf{FysicsWorld} & \textbf{Video-MME} \\
        \midrule

        Unified-IO-2 L~\cite{lu2024unifiedio2} & 1B &
        27.06 & 23.30 & 27.40 & 45.34 & 45.20 \\

        Unified-IO-2 XL~\cite{lu2024unifiedio2} & 3B &
        38.00 & 24.70 & 28.30 & 47.62 & 46.80 \\

        Qwen2.5-Omni-3B~\cite{xu2025qwen25omni} & 3B &
        45.18 & 44.45 & 51.35 & 51.49 & 62.00 \\

        Qwen2.5-Omni-7B~\cite{xu2025qwen25omni} & 7B &
        \underline{56.13} & 45.40 & 53.42 &
        \underline{58.58} & 64.30 \\

        OmniVinci~\cite{ye2025omnivinci} & 7B &
        46.47 & \underline{48.23} & \underline{66.50} & 55.52 & \underline{68.20} \\

        Unified-IO-2 XXL~\cite{lu2024unifiedio2} & 7B &
        33.98 & 25.90 & 28.24 & 47.62 & 54.40 \\

        \rowcolor{gray!10}
        \textbf{OmniFysics-Nano-V2} & 4B &
        \textbf{60.95} & \textbf{60.32} & \textbf{85.24} &
        \textbf{60.87} & \textbf{80.89} \\

        \bottomrule
    \end{tabular}
\end{table*}

\subsection{Experimental Setup}
\noindent\textbf{Model and Training.} We initialize the shared backbone and visual interface from Qwen3.5-4B~\cite{qwen2026qwen35}, the acoustic encoder from Whisper Medium~\cite{radford2023robust}, and the speech-token and codec decoders from CosyVoice3~\cite{du2025cosyvoice3inthewildspeech}. These pretrained components retain their native tokenization and input-processing schemes, while the newly introduced modality projectors and answer-conditioning modules are optimized according to the stage-wise training strategy. Supervised training is implemented with VeOmn~\cite{ma2025veomni}, whereas reinforcement learning is conducted with MS-SWIFT~\cite{zhao2024msswift}. All training stages are performed on 64 NVIDIA H100 GPUs using bfloat16 precision.

\noindent\textbf{Evaluation.} We evaluate five capability groups: general multimodal understanding, audio understanding, omni-modal and video understanding, physical-world understanding and reasoning, and mathematical and physical reasoning, using each benchmark's native metric. We also conduct a human evaluation of generated speech. 

\subsection{Main Results}
OmniFysics-Nano-V2 exhibits its clearest gains on temporal, cross-modal, and physics-oriented evaluation. It ranks first among the listed systems on the representative benchmarks in Figure~\ref{fig2}(b). The largest margins appear on Daily-Omni and PhysUniBench, followed by MMStar and MMAR; the gains on OmniBench, PhysBench, and PhyX are smaller but remain positive. The complete comparisons in Tables~\ref{tab:main_results}--\ref{tab:reasoning_results} and Figure~\ref{fig:audio_understanding_bar} further show that these targeted strengths are achieved while maintaining competitive performance on general-purpose visual-language benchmarks.

\begin{table*}[t]
    \centering
    \small
    \caption{Physical-world understanding and reasoning for vision-language and omni-modal models. FysicsEval is decomposed into prediction, reasoning, and understanding, and is reported alongside PhysBench, PAI-Bench, QuantiPhy, and PhysUniBench. Size is the reported parameter count in billions (B). Higher scores are better. Bold and underlined values denote the best and second-best results, respectively, in each column; the shaded row is our model.}
    \label{tab:physics_results}
    \setlength{\tabcolsep}{2.5pt}
    \begin{tabular}{l c c c c c c c c}
        \toprule
        \multirow{2}{*}{\textbf{Model}} 
        & \multirow{2}{*}{\textbf{Size}} 
        & \multicolumn{3}{c}{\textbf{FysicsEval}} 
        & \multirow{2}{*}{\textbf{PhysBench}} 
        & \multirow{2}{*}{\begin{tabular}[c]{@{}c@{}}\textbf{PAI-Bench}\end{tabular}} 
        & \multirow{2}{*}{\textbf{QuantiPhy}} 
        & \multirow{2}{*}{\begin{tabular}[c]{@{}c@{}}\textbf{PhysUniBench}\end{tabular}} \\
        
        \cmidrule(lr){3-5}
        
        & &
        \textbf{Prediction} &
        \textbf{Reasoning} &
        \textbf{Understanding} &
        & & & \\
        
        \midrule

        Ovis2.5~\cite{lu2025ovis25} & 2B &
        20.40 & 2.46 & 89.50 &
        43.80 & 42.70 & 29.30 & 37.00 \\

        SAIL-VL2~\cite{yin2025sailvl2} & 2B &
        21.90 & \underline{2.58} & 84.70 &
        44.40 & 48.10 & 25.60 & 37.30 \\

        Ovis-U1~\cite{wang2025ovisu1} & 3B &
        6.60 & 2.22 & 81.90 &
        26.40 & 21.20 & 28.60 & 38.10 \\

        Qwen2.5-Omni-3B~\cite{xu2025qwen25omni} & 3B &
        18.10 & 1.71 & 87.50 &
        35.50 & 50.60 & 28.20 & 33.40 \\

        Qwen3.5-4B~\cite{qwen2026qwen35} & 4B &
        24.00 & 2.33 & \underline{92.70} &
        \underline{48.00} & 30.80 & 27.90 & \underline{48.00} \\

        Qwen3-VL-4B-Instruct~\cite{bai2025qwen3vl} & 4B &
        24.80 & 2.24 & 87.70 &
        41.20 & 50.60 & 29.00 & 41.90 \\

        Qwen2.5-Omni-7B~\cite{xu2025qwen25omni} & 7B &
        \underline{27.90} & 2.13 & 86.30 &
        46.30 & \underline{53.00} & \underline{34.40} & 46.40 \\

        OmniVinci~\cite{ye2025omnivinci} & 7B &
        14.30 & 2.07 & 88.30 &
        45.80 & 52.20 & 21.90 & 41.86 \\

        \rowcolor{gray!10}
        \textbf{OmniFysics-Nano-V2} & \textbf{4B} &
        \textbf{45.09} & \textbf{3.26} & \textbf{98.27} &
        \textbf{50.07} & \textbf{54.61} & \textbf{40.82} & \textbf{59.42} \\

        \bottomrule
    \end{tabular}
\end{table*}

\begin{figure}[t]
    \centering
    \small
    \includegraphics[width=0.65\linewidth]{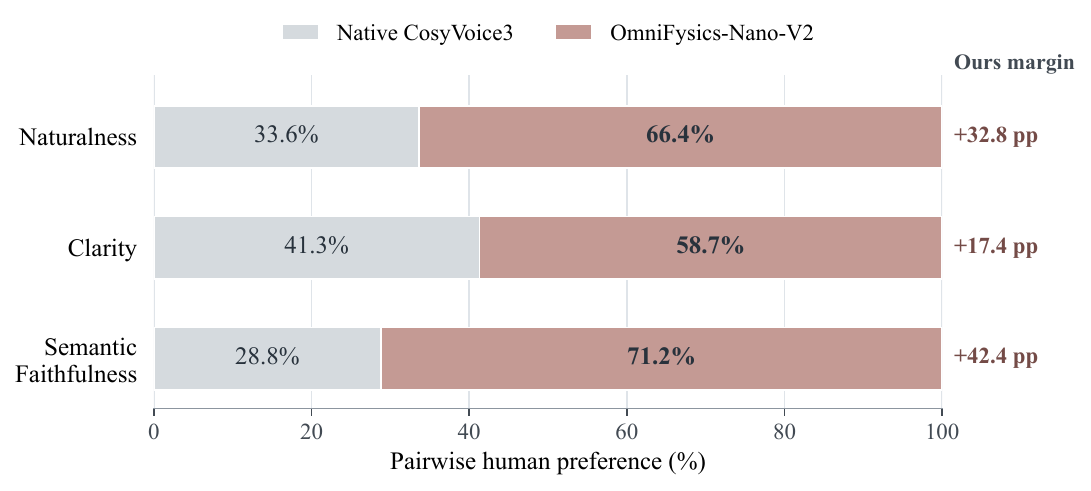}
    \caption{Pairwise human preference (\%) for generated speech from OmniFysics-Nano-V2 and native CosyVoice3, measured by naturalness, clarity, and semantic faithfulness.}
    \label{fig:speech_human_evaluation}
\end{figure}

\subsubsection{General Multimodal Understanding} OmniFysics-Nano-V2 is strongest among the listed omni-modal models on MMStar~\cite{chen2024mmstar}, MMMU~\cite{yue2024mmmu}, and HallusionBench~\cite{guan2024hallusionbench}, with scores of 75.27, 62.11, and 62.49, respectively (Table~\ref{tab:main_results}). Relative to Qwen2.5-Omni-7B, the corresponding gains are 11.27, 2.91, and 17.59 points, together with a 3.73-point gain on MMBench-V1.1~\cite{liu2023mmbench}. OmniVinci remains strongest on MMBench-V1.1 and AI2D, whereas the Qwen3.5-4B reference leads on MMMU. The resulting profile highlights a clear advantage on several omni-modal general-understanding tasks while preserving competitive performance across the broader visual-language suite.

\subsubsection{Audio Understanding} 
Audio-input evaluation likewise highlights the model's parameter efficiency. OmniFysics-Nano-V2 reaches 76.80 on MMAU~\cite{sakshi2025mmau} and 66.10 on MMAR~\cite{ma2025mmar}, exceeding Qwen2.5-Omni-7B by 5.30 and 9.40 points and Kimi-Audio by 8.60 and 27.31 points, respectively (Figure~\ref{fig:audio_understanding_bar}). It therefore provides the strongest audio-understanding results among the listed systems despite having fewer parameters than the 7B baselines, suggesting that the proposed physical-perception training transfers beyond visual inputs.

\subsubsection{Omni-Modal and Video Understanding} The largest improvements emerge on tasks that require temporal or cross-modal evidence integration. OmniFysics-Nano-V2 ranks first among the listed models on all five benchmarks in Table~\ref{tab:omni_results}, surpassing Qwen2.5-Omni-7B on OmniBench~\cite{li2024omnibench}, WorldSense~\cite{hong2026worldsense}, Daily-Omni~\cite{zhou2026dailyomni}, FysicsWorld~\cite{jiang2025fysicsworld}, and Video-MME~\cite{fu2025videomme} by 4.82, 14.92, 31.82, 2.29, and 16.59 points, respectively. The particularly large gains on Daily-Omni, Video-MME, and WorldSense align with the visual--acoustic integration targeted by the model.

\subsubsection{Physics Understanding and Reasoning} Physics-oriented evaluation provides the most direct test of the proposed supervision, and the advantage is consistent across task types. OmniFysics-Nano-V2 obtains 45.09, 3.26, and 98.27 on FysicsEval~\cite{han2026omnifysicsphysicalintelligenceevolution} Prediction, Reasoning, and Understanding, improving over Qwen2.5-Omni-7B by 17.19, 1.13, and 11.97 points (Table~\ref{tab:physics_results}). It also leads all listed models on PhysBench~\cite{chow2025physbench}, PAI-Bench~\cite{zhou2026paibench}, QuantiPhy~\cite{li2025quantiphy}, and PhysUniBench~\cite{wang2025physunibench}; relative to the same 7B baseline, the gains are 3.77, 1.61, 6.42, and 13.02 points. These gains span physical properties, perception, and grounded reasoning, indicating that the proposed supervision benefits multiple forms of physical-world evaluation.
\subsubsection{Mathematical and Physical Reasoning} Beyond perception-focused benchmarks, OmniFysics-Nano-V2 retains an advantage on reasoning-intensive tasks. It reaches 13.33 on AIME25~\cite{aime25}, 28.13 on TheoremQA~\cite{chen2023theoremqa}, and 48.63 on PhyX~\cite{shen2025phyx} (Table~\ref{tab:reasoning_results}). Relative to Qwen2.5-Omni-7B, the gains on TheoremQA and PhyX are 2.63 and 1.96 points, while the AIME25 score increases from 10.00 to 13.33. The strongest improvements therefore occur on the theorem-level and physically grounded tasks that align with the proposed training and evaluation design.

\begin{table}[htbp]
    \centering
    \small
    \caption{Mathematical and physically grounded reasoning results on AIME25, AIME26, TheoremQA, and PhyX. Model size is reported in billions of parameters (B), and higher scores indicate better performance. A ``-'' indicates that the model did not produce a valid response on the corresponding benchmark; hence, no score is reported. Bold and underlined values denote the best and second-best distinct results, respectively, with ties receiving the same formatting. The shaded row highlights our model.}
    \label{tab:reasoning_results}
    \setlength{\tabcolsep}{3pt}
    \begin{tabular}{l c c c c c}
        \toprule
        \textbf{Model} & \textbf{Size} & \textbf{AIME25} &
        \textbf{AIME26} & \textbf{TheoremQA} & \textbf{PhyX} \\
        \midrule

        Qwen2.5-Omni-3B\cite{xu2025qwen25omni} & 3B &
        - & \underline{3.33} & 20.13 & 32.33 \\

        Qwen2.5-Omni-7B\cite{xu2025qwen25omni} & 7B &
        \underline{10.00} & \textbf{10.00} &
        \underline{25.50} & \underline{46.67} \\

        Baichuan-Omni-1.5 \cite{li2025baichuan} & 7B &
        - & - & 14.13 & 37.40 \\

        OmniVinci\cite{ye2025omnivinci} & 7B &
        - & - & 12.00 & 35.23 \\

        MiniCPM-o-4\_5\cite{cui2026minicpmo45} & 9B &
        3.33 & \underline{3.33} &
        23.38 & 32.17 \\

        \rowcolor{gray!10}
        \textbf{OmniFysics-Nano-V2} & 4B &
        \textbf{13.33} & \underline{3.33} &
        \textbf{28.13} & \textbf{48.63} \\

        \bottomrule
    \end{tabular}
\end{table}

\subsubsection{Speech Output Quality}

The advantage extends from input understanding to generated speech. In the paired human evaluation shown in Figure~\ref{fig:speech_human_evaluation}, OmniFysics-Nano-V2 is preferred to native CosyVoice3 on all three criteria, with margins of 32.8 percentage points for naturalness, 17.4 for clarity, and 42.4 for semantic faithfulness. The largest margin occurs in semantic faithfulness, showing that the answer-conditioned speech path preserves the content of the multimodal response while also improving perceived naturalness and clarity.

\subsection{Ablation Study}
\label{sec:ablation_study}
\subsubsection{Static and Dynamic Physical Supervision}
As shown in Table~\ref{tab:static_dynamic_data_ablation}, we compare Full SFT with variants that remove Static Physical Data, Dynamic Physical Data, or both. All variants share the same initialization, model, optimizer, and epoch-level training recipe, and are evaluated before policy optimization.

\newcommand{\sftrowbg}{%
  \makebox[0pt][l]{%
    \smash{%
      \raisebox{-0.3\baselineskip}{%
        \color{gray!10}\rule{\textwidth}{0.9\baselineskip}%
      }%
    }%
  }%
}

\begin{table*}[t]
\centering
\caption{Ablation of static and dynamic physical data. Checkmarks indicate the included data branch. All variants use the same initialization and training recipe and are evaluated after SFT and before RL. Bold and \underline{underlined} values denote the best and second-best results, respectively.}
\label{tab:static_dynamic_data_ablation}

\small
\renewcommand{\arraystretch}{1.0}
\setlength{\tabcolsep}{2.5pt}

\begin{tabular*}{\textwidth}{
@{}
>{\raggedright\arraybackslash}p{0.275\textwidth}
@{\hspace{5pt}}
>{\centering\arraybackslash}p{0.057\textwidth}
@{\hspace{2pt}}
>{\centering\arraybackslash}p{0.066\textwidth}
@{\hspace{6pt}}
@{\extracolsep{\fill}}
ccccc
@{}
}
\toprule

\multirow{2}{*}[-2.5pt]{\textbf{Training Variant}} &
\multirow{2}{*}[-2.5pt]{\textbf{Static}} &
\multirow{2}{*}[-2.5pt]{\textbf{Dynamic}} &
\multicolumn{3}{c}{\textbf{Physics Und.}} &
\multicolumn{2}{c}{\textbf{General Multimodal Und.}} \\

\cmidrule(lr){4-6}
\cmidrule(lr){7-8}

& & &
\textbf{PhysBench} &
\textbf{QuantiPhy} &
\textbf{PhysUniBench} &
\textbf{MMBench-V1.1} &
\textbf{MMStar} \\

\midrule
\textit{w/o} Static \& Dynamic Physical Data
&
&
& 45.98
& 30.20
& 49.20
& \underline{81.91}
& 72.21 \\

\textit{w/o} Static Physical Data
&
& \checkmark
& \underline{46.64}
& 31.25
& \underline{55.79}
& 81.90
& \underline{72.53} \\

\textit{w/o} Dynamic Physical Data
& \checkmark
&
& 46.53
& \underline{37.10}
& 55.53
& 81.87
& 72.45 \\

\sftrowbg\raisebox{-1.5pt}{\textbf{Full SFT}}
& \raisebox{-1.5pt}{\checkmark}
& \raisebox{-1.5pt}{\checkmark}
& \raisebox{-1.5pt}{\textbf{46.72}}
& \raisebox{-1.5pt}{\textbf{38.72}}
& \raisebox{-1.5pt}{\textbf{56.67}}
& \raisebox{-1.5pt}{\textbf{82.40}}
& \raisebox{-1.5pt}{\textbf{72.82}} \\

\midrule
\midrule
\end{tabular*}

\vspace{-0.35em}

\begin{tabular*}{\textwidth}{
@{}
>{\raggedright\arraybackslash}p{0.275\textwidth}
@{\hspace{5pt}}
>{\centering\arraybackslash}p{0.057\textwidth}
@{\hspace{2pt}}
>{\centering\arraybackslash}p{0.066\textwidth}
@{\extracolsep{\fill}}
cccc
@{}
}

\multirow{2}{*}[-2.5pt]{\textbf{Training Variant}} &
\multirow{2}{*}[-2.5pt]{\textbf{Static}} &
\multirow{2}{*}[-2.5pt]{\textbf{Dynamic}} &
\multicolumn{3}{c}{\textbf{Omni-Modal Und.}} &
\multicolumn{1}{c}{\textbf{Audio Und.}} \\

\cmidrule(lr){4-6}
\cmidrule(lr){7-7}

& & &
\textbf{FysicsWorld} &
\textbf{OmniBench} &
\textbf{Daily-Omni} &
\textbf{MMAR} \\

\midrule
\textit{w/o} Static \& Dynamic Physical Data
&
&
& 50.85
& 49.76
& 72.45
& 57.62 \\

\textit{w/o} Static Physical Data
&
& \checkmark
& \underline{57.61}
& \underline{51.28}
& \underline{81.61}
& \underline{59.10} \\

\textit{w/o} Dynamic Physical Data
& \checkmark
&
& 51.42
& 50.31
& 73.18
& 57.50 \\

\sftrowbg\raisebox{-1.5pt}{\textbf{Full SFT}}
& \raisebox{-1.5pt}{\checkmark}
& \raisebox{-1.5pt}{\checkmark}
& \raisebox{-1.5pt}{\textbf{58.45}}
& \raisebox{-1.5pt}{\textbf{52.54}}
& \raisebox{-1.5pt}{\textbf{82.58}}
& \raisebox{-1.5pt}{\textbf{59.80}} \\
\bottomrule
\end{tabular*}
\end{table*}

The two supervision branches make complementary contributions. Full SFT is the strongest variant on all nine reported benchmarks, reaching 46.72 on PhysBench, 38.72 on QuantiPhy, 56.67 on PhysUniBench, 58.45 on FysicsWorld, 52.54 on OmniBench, 82.58 on Daily-Omni, 59.80 on MMAR, 82.40 on MMBench-V1.1, and 72.82 on MMStar; relative to removing both branches, these results improve by 0.74, 8.52, 7.47, 7.60, 2.78, 10.13, 2.18, 0.49, and 0.61 points, respectively. Static supervision has its clearest effect on property-oriented evaluation: removing it lowers QuantiPhy by 7.47 points and also reduces PhysUniBench, FysicsWorld, OmniBench, and Daily-Omni by 0.88, 0.84, 1.26, and 0.97 points, respectively. Dynamic supervision is more important for temporal and cross-modal behavior, as removing it lowers FysicsWorld by 7.03 points, OmniBench by 2.23 points, and Daily-Omni by 9.40 points. It is also the primary contributor to MMAR: removing dynamic supervision lowers the score by 2.30 points, whereas removing static supervision lowers it by 0.70 points. Relative to the variant without either branch, dynamic supervision alone improves MMAR by 1.48 points, while static supervision alone changes it by $-0.12$ points; combining both branches yields the best score of 59.80. This result indicates that static supervision does not benefit MMAR in isolation but provides a complementary gain when paired with dynamic supervision, consistent with the combined model's strongest overall physical and omni-modal profile.

\subsubsection{Joint Audio--Visual Understanding}
Understanding the physical world requires audio and visual information to be perceived and interpreted jointly; either modality alone provides only a partial account of real-world events. As shown in Table~\ref{tab:modality_ablation}, joint audio--visual input outperforms both single-modality settings for every model and benchmark. For OmniFysics-Nano-V2, retaining both streams improves over audio-only input by 15.50--33.50 points and over vision-only input by 14.23--40.42 points across the five benchmarks. The stronger isolated modality also varies by task: audio is more informative on OmniBench and FysicsWorld, whereas vision is stronger on WorldSense, Daily-Omni, and Video-MME. This task-dependent pattern reflects the distinction raised in the Introduction: vision captures objects, geometry, and motion, while audio provides event timing, contact, and material-response cues. These observations must be understood together to connect what happens, when it happens, and the physical cause behind it. The consistent degradation after removing either stream therefore confirms that simultaneous audio--visual understanding is essential for understanding the physical world.

\begin{table*}[t]
    \centering
    \small
    \caption{Missing-modality evaluation on omni-modal benchmarks. A, V, and A+V retain the audio stream, visual stream, and both streams, respectively; the textual question is unchanged. Higher scores are better. Rows for our model are shaded, and bold values denote its full-input results.}
    \label{tab:modality_ablation}
    \setlength{\tabcolsep}{6pt}
    \renewcommand{\arraystretch}{0.8}
    \begin{tabular}{lcccccc}
        \toprule
        \textbf{Model} & \textbf{Input} & \textbf{OmniBench} & \textbf{WorldSense} &
        \textbf{Daily-Omni} & \textbf{FysicsWorld} & \textbf{Video-MME} \\
        \midrule
        \multirow{3}{*}{Qwen2.5-Omni-3B~\cite{xu2025qwen25omni}}
        & A   & 33.01 & 40.05 & 49.12 & 27.66 & 49.85 \\
        & V   & 34.15 & 33.76 & 39.79 & 23.57 & 49.67 \\
        & A+V & 45.18 & 44.45 & 51.35 & 51.49 & 62.00 \\
        \midrule
        \multirow{3}{*}{Qwen2.5-Omni-7B~\cite{xu2025qwen25omni}}
        & A   & 32.49 & 41.95 & 50.14 & 29.71 & 52.37 \\
        & V   & 36.51 & 34.21 & 41.88 & 24.38 & 53.52 \\
        & A+V & 56.13 & 45.40 & 53.42 & 58.58 & 64.30 \\
        \midrule
        \multirow{3}{*}{OmniVinci~\cite{ye2025omnivinci}}
        & A   & 34.94 & 43.08 & 51.61 & 28.80 & 57.26 \\
        & V   & 37.39 & 41.80 & 52.79 & 26.33 & 62.85 \\
        & A+V & 46.47 & 48.23 & 66.50 & 55.52 & 68.20 \\
        \midrule
        \rowcolor{gray!10}
        & A   & 45.45 & 42.55 & 54.34 & 27.37 & 54.78 \\
        \rowcolor{gray!10}
        \textbf{OmniFysics-Nano-V2} & V   & 43.61 & 46.09 & 55.21 & 20.45 & 63.30 \\
        \rowcolor{gray!10}
        & A+V & \textbf{60.95} & \textbf{60.32} & \textbf{85.24} & \textbf{60.87} & \textbf{80.89} \\
        \bottomrule
    \end{tabular}
\end{table*}

\begin{figure*}[t]
    \centering
    \includegraphics[width=1.0\linewidth]{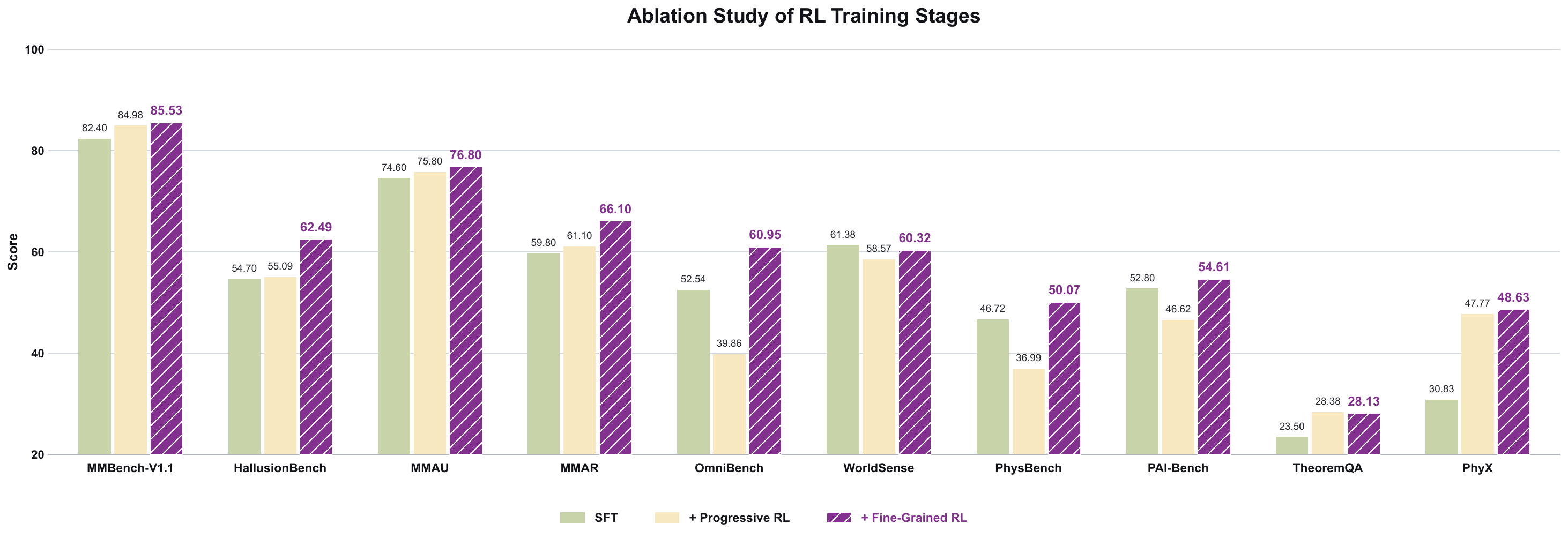}
    \caption{Successive checkpoints across ten understanding and reasoning benchmarks. SFT denotes the Stage~1-3 checkpoint; Progressive RL adds Stage~2-1 answer-correctness optimization; Fine-Grained RL further adds Stage~2-2 physical-perception and format rewards. Each row continues training from the preceding row. Higher scores are better.}
    \label{fig:AblationofRL}
\end{figure*}




\begin{figure*}[t]
    \centering
    \includegraphics[width=1.0\linewidth]{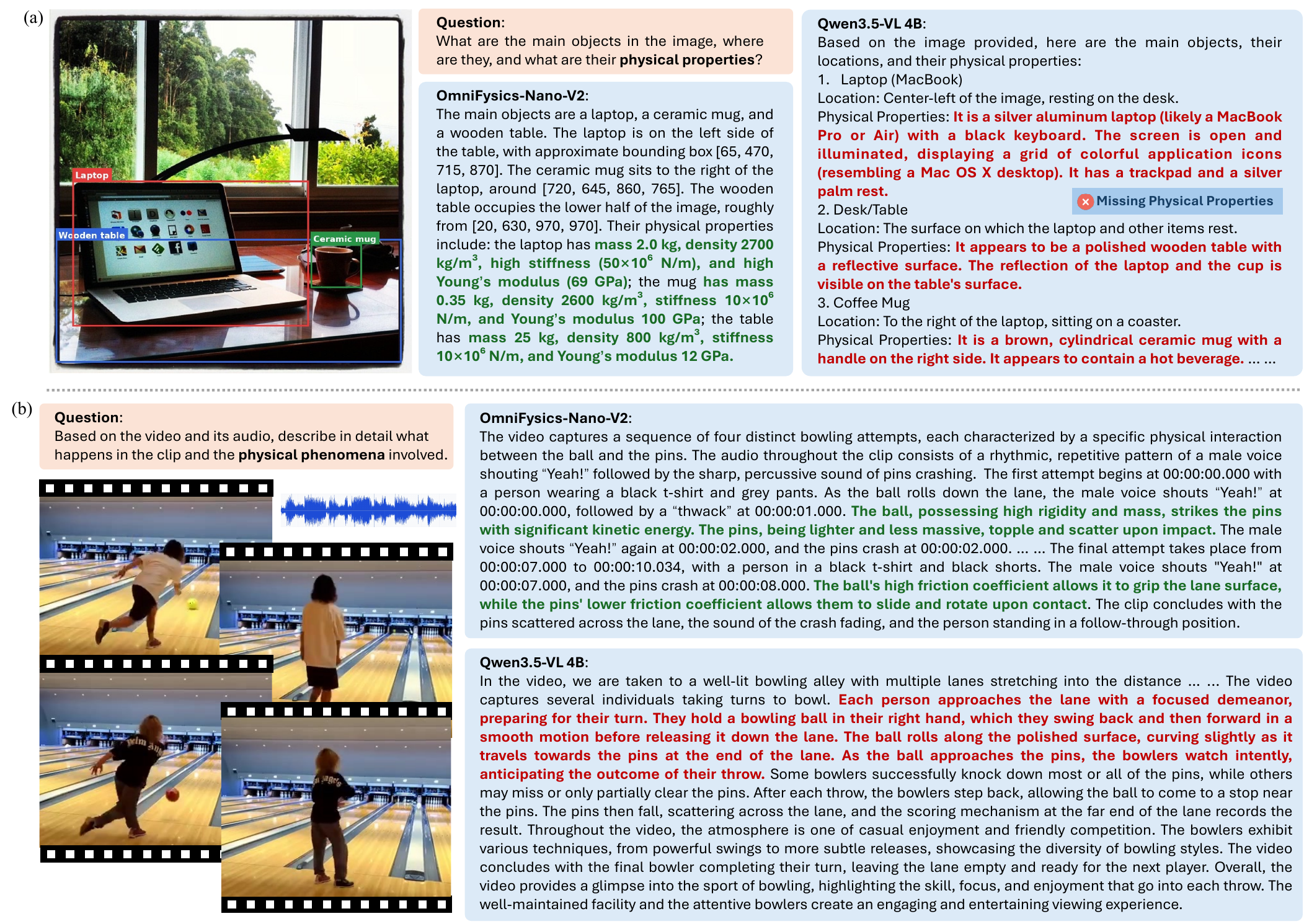}
    \caption{Qualitative case studies of physical property prediction and multimodal physical understanding.}
    \label{fig:case_study}
\end{figure*}
\begin{table}[t]
    \centering
    \small
    \caption{Reward-diversity filtering across four RL data categories. Reduction rate is the fraction of candidate samples removed, and saved GPU-hours is the estimated avoided training compute. The estimates total 2,687.7 GPU-hours.}
    \label{tab:data_reduction_gpu_savings}
    \setlength{\tabcolsep}{3.5pt}
    \begin{tabular}{lcc}
        \toprule
        \textbf{Training Data Type}
        & \textbf{Reduction Rate}
        & \textbf{Saved GPU-Hours} \\
        \midrule
        Mathematical Reasoning
        & 73.99\%
        & 1,239 \\

        Image-Text Perception
        & 44.15\%
        & 416.3 \\

        Video Perception
        & 69.79\%
        & 530.2 \\

        Multimodal Perception
        & 57.69\%
        & 502.2 \\
        \bottomrule
    \end{tabular}
\end{table}

\subsubsection{Progressive Reinforcement Learning}
Figure~\ref{fig:AblationofRL} reports the successive checkpoints obtained after SFT, Progressive Multimodal RL, and Fine-Grained RL. Each row continues from the preceding checkpoint, allowing the two policy-optimization stages to be examined separately. Relative to SFT, the final checkpoint improves nine of the ten benchmarks. The gains are largest on PhyX (+17.80 points), OmniBench (+8.41), and HallusionBench (+7.79), with additional improvements on MMBench-V1.1, MMAU, MMAR, PhysBench, PAI-Bench, and TheoremQA. WorldSense changes by only -1.06 points, indicating that the final policy retains its overall level of temporal understanding while improving the other reported capabilities.

The two stages exhibit distinct effects. Progressive Multimodal RL raises TheoremQA and PhyX by 4.88 and 16.94 points over SFT, while the intermediate checkpoint is lower on OmniBench and PhysBench. Fine-Grained RL subsequently increases OmniBench, WorldSense, PhysBench, and PAI-Bench by 21.09, 1.75, 13.08, and 7.99 points relative to the intermediate checkpoint, and further improves HallusionBench and MMAR by 7.40 and 5.00 points. TheoremQA decreases by only 0.25 points in this stage and remains 4.63 points above SFT. These results support a staged optimization strategy in which answer-level policy improvement is followed by intermediate physical-perception feedback to recover cross-modal performance and strengthen physical reasoning.

\subsubsection{Reward-Diversity Filtering}
Reward-diversity filtering substantially reduces the RL corpus while retaining prompts that provide a non-degenerate group-relative signal. For each candidate prompt, four stochastic responses are sampled and scored; a prompt is retained only when its rollout group contains more than one attained reward value. Figure~\ref{fig:rl_data} illustrates this criterion, and Table~\ref{tab:data_reduction_gpu_savings} reports its category-level effect. The filter removes 73.99\% of Mathematical Reasoning, 44.15\% of Image--Text Perception, 69.79\% of Video Perception, and 57.69\% of Multimodal Perception samples. These reductions correspond to estimated savings of 1,239, 416.3, 530.2, and 502.2 GPU-hours, respectively, or 2,687.7 GPU-hours in total. The results show that the filtering rule removes groups that cannot provide a relative reward ordering and concentrates policy optimization on prompts for which the sampled responses expose an observable quality difference.

\begin{figure*}[t]
    \centering
    \includegraphics[width=1.0\linewidth]{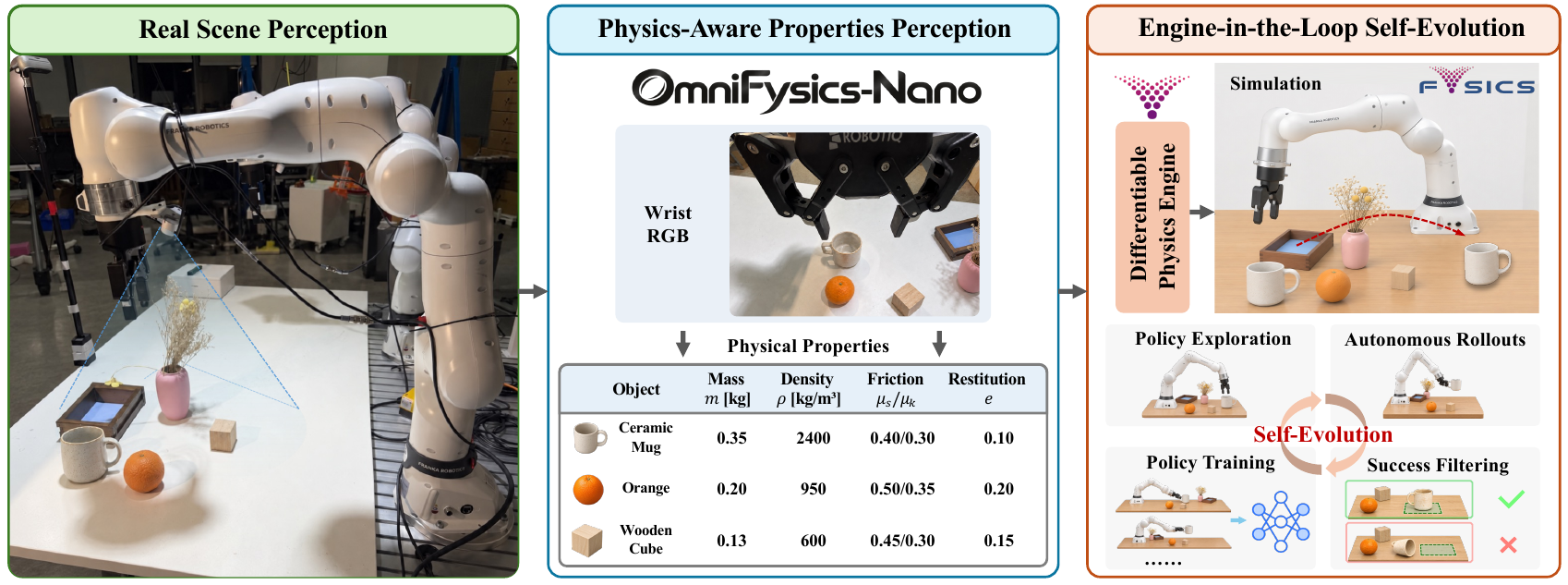}
    \caption{The proposed OmniFysics-Nano-V2 is integrated with our in-house differentiable physics engine, enabling continual self-evolution of the embodied agent.}
    \label{fig:application}
\end{figure*}

\subsection{Case Study}
To illustrate the behavior underlying the aggregate results, Figure~\ref{fig:case_study} compares a static object--property query with a bowling clip accompanied by its soundtrack. In Figure~\ref{fig:case_study}(a), both models identify the laptop, mug, and table, whereas OmniFysics-Nano-V2 additionally provides object-level boxes and a structured profile of mass, density, stiffness, and Young's modulus. These values are prior-informed estimates rather than measurements; the example therefore demonstrates grounded output structure, while quantitative reliability is evaluated by the physics benchmarks. In Figure~\ref{fig:case_study}(b), both models recover the visible bowling sequence, but OmniFysics-Nano-V2 also associates vocalizations and pin-crash sounds with the interaction stages and explains the outcome using kinetic-energy transfer, friction, sliding, rotation, and mass differences. This audio--visual evidence chain is consistent with Table~\ref{tab:modality_ablation}, where A+V outperforms either single modality on every reported benchmark.

\section{Application in Embodied Scenarios}
Beyond offline benchmark evaluation, we further integrate OmniFysics-Nano-V2 into an embodied manipulation pipeline that connects scene perception, physics-aware property estimation, and engine-in-the-loop policy optimization. Figure~\ref{fig:application} shows the progress.
The robot first perceives the objects and their spatial configuration in the workspace, while its wrist-mounted RGB camera provides a close-range view from the manipulation perspective.
Taking only the wrist-camera RGB image as input, OmniFysics-Nano-V2 predicts object-level physical properties for the visible objects, including mass, density, static and kinetic friction coefficients, and restitution.
These predictions convert visual appearance into structured physical priors and, together with the perceived object configuration, are used to parameterize the corresponding object instances in the physics engine.
This process produces a task-specific simulation environment in which object motion and contact evolve according to the predicted dynamics and contact parameters.
OmniFysics-Nano-V2 therefore does not directly generate robot actions, but instead serves as an interface between visual scene perception and simulation-ready physical states.
Within the parameterized environment, the policy first explores candidate interactions and then performs autonomous rollouts to evaluate different actions under the estimated object properties and contact conditions.
Each rollout is filtered according to task success: successful trajectories are retained as valid training samples, whereas failed trajectories are excluded from the current training set.
The filtered successful trajectories are subsequently used to train and update the policy, which then returns to the simulation environment for another round of exploration and autonomous rollout.
This recurring cycle of physics-engine simulation, policy exploration, autonomous rollout, success filtering, and policy training constitutes an engine-in-the-loop policy self-evolution process.
By providing object-specific physical conditions for policy optimization, this pipeline reduces reliance on repeated trial and error with the physical robot and turns the model's physical understanding into actionable support for contact-rich manipulation and scalable embodied-data construction.

\section{Conclusion}
We presented OmniFysics-Nano-V2, a compact omni-modal model for physical-world understanding that unifies image, video, audio, speech, and text within a shared reasoning framework and supports answer-conditioned speech generation. 
The proposed approach combines static object--property supervision, dynamic audio--visual event supervision, reward-diversity filtering, and progressive policy optimization. 
Experiments show that these components improve complementary aspects of physical intelligence: static supervision supports property-oriented reasoning, dynamic supervision strengthens temporal and cross-modal understanding, and joint audio--visual perception provides more complete evidence for physical interactions. 
The resulting model achieves strong performance across omni-modal, video, physical-world, and reasoning benchmarks while retaining general multimodal competence, demonstrating the value of explicitly supervising the evidence chain from physical observations to grounded answers.
Beyond offline evaluation, we further explore coupling OmniFysics-Nano-V2 with a differentiable physics engine, where object-level property estimates parameterize simulation and are refined according to discrepancies between simulated and observed trajectories. 
This model--engine interaction provides a framework for connecting omni-modal physical perception with physically grounded state prediction and embodied policy optimization. 
By equipping AI systems with both omni-modal and physical-world perception capabilities, we believe that OmniFysics-Nano-V2 has the potential to serve as a cornerstone of next-generation Physical AI, enabling agents to perceive, simulate, reason, and reliably interact with the real world.

\clearpage

\bibliographystyle{plainnat}
\bibliography{main}

\end{document}